\documentclass[journal,transmag]{IEEEtran}

\usepackage{moreverb,url}
\usepackage{array}
\usepackage{multirow}
\usepackage{mdwlist}
\usepackage{graphics}
\usepackage{graphicx}
\usepackage{mathptmx}
\usepackage{amsmath}
\usepackage{amsfonts}
\usepackage{amssymb}
\usepackage[noadjust]{cite}

\begin{document}

\title{Imprecise dynamic walking with time-projection control}


\author{\IEEEauthorblockN{
Salman Faraji\IEEEauthorrefmark{1},
Philippe M\"{u}llhaupt\IEEEauthorrefmark{1}, and
Auke J. Ijspeert\IEEEauthorrefmark{1}}
\IEEEauthorblockA{\IEEEauthorrefmark{1}EPFL, Switzerland}
\thanks{Corresponding author: S. Faraji (email: salman.faraji@epfl.ch).}}

\markboth{ }%
{Shell \MakeLowercase{\textit{et al.}}: Bare Demo of IEEEtran.cls for IEEE Transactions on Magnetics Journals}

\IEEEtitleabstractindextext{%
\begin{abstract}
We present a new walking foot-placement controller based on 3LP, a 3D model of bipedal walking that is composed of three pendulums to simulate falling, swing and torso dynamics. Taking advantage of linear equations and closed-form solutions of the 3LP model, our proposed controller projects intermediate states of the biped back to the beginning of the phase for which a discrete LQR controller is designed. After the projection, a proper control policy is generated by this LQR controller and used at the intermediate time. This control paradigm reacts to disturbances immediately and includes rules to account for swing dynamics and leg-retraction. We apply it to a simulated Atlas robot in position-control, always commanded to perform in-place walking. The stance hip joint in our robot keeps the torso upright to let the robot naturally fall, and the swing hip joint tracks the desired footstep location. Combined with simple Center of Pressure (CoP) damping rules in the low-level controller, our foot-placement enables the robot to recover from strong pushes and produce periodic walking gaits when subject to persistent sources of disturbance, externally or internally. These gaits are imprecise, i.e., emergent from asymmetry sources rather than precisely imposing a desired velocity to the robot. Also in extreme conditions, restricting linearity assumptions of the 3LP model are often violated, but the system remains robust in our simulations. An extensive analysis of closed-loop eigenvalues, viable regions and sensitivity to push timings further demonstrate the strengths of our simple controller.
\end{abstract}

}

\maketitle
\IEEEpeerreviewmaketitle

\section{Introduction}
\label{sec::intro}

Bipedal locomotion is a challenging task for humanoid robots due to actuation complexities, the large number of degrees of freedom and most importantly, the hybrid nature of changing contacts. The robot should keep upright balance, bear its weight with the lower-limbs and at the same time, track desired trajectories very quickly to ensure dynamic stability. Contact changes also require physical or virtual compliance in the legs for unforeseen touch-down events. In this paper, we shortly review dynamic walking and discuss two possible ways of stabilizing the motion: using ankle torques and foot-placement. We then propose a new theory and a very simple controller that combines these strategies and achieves dynamic walking at different gait parameters. This controller eliminates the need for calculating reference walking trajectories and thus produces a high-level compliant behavior. The same controller is also used in extreme push-recovery scenarios which are, in principle, similar to our walking scenarios. In all these cases, behaviors emerge from internal or external sources of asymmetry which continuously interact with our controller. 

\subsection{Center of Pressure Control}
A conventional way of stabilizing walking is to use ankle torques for tracking Center of Mass (CoM) trajectories. In this method, dynamically consistent motion plans are generated via simplified models and replicated on the robot through active control of the Center of Pressure (CoP). Such tracking can use contact force sensors \cite{kajita1991study} or inverse dynamics and torque control \cite{kuindersma2016optimization, feng20133d}. In this paradigm, the hybrid nature of walking does not limit the controller, since ankle torques can stabilize the trajectories immediately. Therefore, simple PID gains \cite{feng20133d} or continuous-time LQR approaches \cite{kuindersma2014efficiently} can easily correct for CoM deviations. With the CoP control strategy, it is also possible to generate more human-like motions with prescribed toe-off and heel-contact phases. However, at least one foot should always be in full contact with the terrain to allow for CoP modulation \cite{ogura2006human, griffin2018straight, miura2011human}. In summary, kinematic trajectories do not change much in this control approach, and only the ankle torques resist against perturbations and deviations. 

\subsection{Foot-Placement}
Another method of walking stabilization relies on foot-placement \cite{herdt2010walking, faraji2014robust, capturability, feng2016robust} which assumes no ankle torques available in sagittal and lateral directions (a foot segment may still be needed in case of 3D walking to provide transversal contact wrenches). In this strategy, we assume under-actuation in the system during each single-support phase, i.e., the robot always falls in some direction. Foot-placement provides stability over the next phases of motion where the adjusted foot location leads to capturing or pumping extra energy in the system and bringing it back to the nominal trajectory. Once the adjustment is calculated with an advanced strategy, we can use PID or continuous LQR controllers on the swing-hip joint to reach the new target-point. The hybrid nature of walking requires prediction of next motion phases which makes foot-placement challenging. However, simplified models of the robot can speed up calculations considerably. Inverted Pendulum (IP) \cite{kuo2005energetic} and its linear version (LIP) \cite{kajita1991study} are probably the simplest models used for these predictions, concentrating the whole mass of the robot into a point and modeling the legs with massless inverted pendulums.

\subsection{Discrete Control}
With simplified models we can discretize the motion, i.e., to decrease our control resolution and think about trajectories (and deviations) only at specific phase events. Such discretization, e.g., between touch down or maximum apex moments, can form the basis for a discrete controller that adjusts inputs when the event triggers \cite{rummel2010stable}. The LIP model provides analytical transition matrices in this regard while nonlinear models require numerical integrations to obtain similar Poincar\'e maps \cite{poincare}. They also need a library of controllers to handle such linearization around different gait conditions \cite{kelly2015non, manchester2014real, gregg2012control}. The synchrony of control with hybrid phase-changes is of particular interest in foot-placement because the correction only becomes effective at the hybrid phase-change moments \cite{zaytsev2015two, kelly2015non, byl2008approximate}. We note that in single-mass models like IP and LIP, we decide the target swing location instantly while in reality, the hip controller requires time to reach these positions. A discrete controller triggered at the maximum apex moment can, therefore, decide a swing attack angle and let the hip joint simply reach it in the rest of the phase. 

\subsection{Continuous Control}
Although a discrete controller can predict the future very rapidly in terms of computation, there might be intermittent disturbances that shortly act on the system at any time and disappear. As mentioned earlier, discretization of walking trajectories is usually synchronized with the gait frequency. However, due to unstable falling dynamics, the stance period (i.e., stance period) is long enough for even moderate intermittent disturbances to accumulate and result in a considerable deviation. Such an unstable nature requires taking a large step in the next phase whereas the footstep location of that same phase could be continuously adjusted online to stabilize with less effort. This raises an interesting control problem in which the discretization rate is reasonable but too slow to handle inter-sample disturbances. In other words, although the next footstep location only becomes effective when established, we need to adjust it online according to continuous deviations we observe at every control tick (e.g., millisecond).

\subsection{Model Predictive Control}
Inter-sample disturbances become important in hybrid systems with switching modes, where deviation of one state (e.g., swing location) might not be important in one mode due to a weak coupling, but becomes very important in another mode where it has a strong coupling (e.g., when it becomes the new stance foot location). In normal continuous systems without hybrid modes, however, the coupling between variables remains the same, and it is possible to discretize at any rate or even apply standard controllers such as Linear Quadratic Regulators (LQR) \cite{ogata1995discrete}. These controllers have extensive applications in CoP control \cite{stephens2007integral, tedrake2015closed}, though, for foot-placement, they are not straightforward and lead to a complicated discrete and switched LQR problem \cite{zhang2009value}. The switching modes here refer to integer decision variables which decided phase changes. These variables make optimizations computationally very expensive \cite{zhang2009value}, though they can lead to a variable timing in walking control which is more powerful \cite{aftab2012ankle} (depending on sampling time). We can remove these decision variables by fixing the rhythm of walking and make the optimization easier to solve, but the hybrid nature should still be considered. Following this idea in an earlier work, we optimized future footstep locations by such a fixed-time Model Predictive Control (MPC) problem on the LIP model \cite{faraji2014robust}. Our optimization was solved every millisecond, and its prediction horizon involved the remaining time of the current phase plus a couple of next phases, similar to \cite{feng2016robust}.

\subsection{Intuition}
Apart from complications of predictive controllers discussed earlier, foot-placement, in the end, translates to a mapping between measured errors and footstep adjustments. Raibert \cite{raibert1984experiments} used a known yet very simple approach for hopping where the footstep adjustment was a function of CoM horizontal speed. A similar rule was used for walking in the SIMBICON controller \cite{simbicon}. These intuitive laws with hand-tuned coefficients moved the footstep location further when a faster forward speed was detected. The idea was formally applied to the LIP model in \cite{capturability} where those coefficients precisely captured the motion, i.e., they found a footstep location on top of which the CoM stops with zero velocity. In both frameworks, the footstep location can be adjusted online in reaction to inter-phase disturbances. The LIP model provides closed-form prediction of the future, allowing the capturability framework to consider physical limitations and calculate capture regions for the robot \cite{capturability}.

\subsection{Time-Projection}
In this work, instead of future prediction, we map the inter-sample states back in time to the beginning of the phase. At each control tick, the measured deviations (caused by inter-sample disturbances) are equivalent to some larger initial-phase deviations that a discrete controller has stabilized by applying an input at the beginning of the phase. This input could be parameters of a piece-wise linear swing-hip torque profile during the phase for example. In that imaginary equivalent system whose evolution passes through the current state of the real system, the input remains constant until the end of the phase, possibly because of zero-order hold assumptions in discretization. The proposed control framework in this paper takes that input and applies it to the real system at the current time. In the next control tick again, the backward mapping is repeated, new equivalent initial states and constant inputs are found, and the real system's input is updated. Under certain conditions (on the discrete controller), this process can stabilize controllable linear systems even-though discretization with a fine resolution also works for them. For hybrid walking systems, however, this process makes more sense because we can update actuator inputs with a high frequency and yet, the hybrid nature of the system imposes a much coarser discretization.

\subsection{Comparison with the Literature}
We use the 3LP model instead of LIP \cite{capturability2, faraji2014robust, feng2016robust} which is a more complete version with three masses that account for falling, swing and torso-balancing dynamics \cite{faraji20173lp}. Thanks to linearity, 3LP provides the necessary transition matrices for backward mapping in time-projection control. By disabling the ankle torques in 3LP, we only focus on foot-placement strategy, and thus, the inputs to our system are swing-hip torques. The state vector also includes pelvis and swing foot positions together. Time-projection control, therefore, gives swing-hip-torque profiles as a function of deviations in the pelvis and swing foot trajectories. We integrate the current state with these profiles to the end of the phase in the imaginary equivalent system and obtain the final footstep adjustment as a function of current deviations. Time-projection provides look-up-tables which are much faster to calculate than numerical MPC optimizations, though with the cost of ignoring inequality constraints. This compromise seems negligible, however, and is discussed in the appendix through a viability analysis \cite{zaytsev2015two}. Note that the currently observed deviation could be attributed to an equivalent average CoP point (in the LIP model) as well which could be used for better capture point prediction \cite{capturability}. CoP serves as input for the LIP model of \cite{capturability} whereas, in our time-projection on the foot-less 3LP model, we find initial deviated states and stabilizing hip torques as if we imitate a perturbed but stable imaginary system. If we switch to the LIP model and CoP control strategy, time-projection can also find stabilizing CoP points, though a simple PID controller may do the same job.

\subsection{Walking Simulations}
Our overall walking controller on the simulated Atlas robot uses known strategies from the literature and only adds our particular foot-placement on top. More generally, the proposed method applies to bipedal walking control on humanoid robots with at least position control capabilities, contact force sensors and an IMU unit on the pelvis. Simple damping rules in the ankles implement CoP control \cite{coros2010generalized}, stance-hip joints keep the torso upright \cite{virtual_model_pratt}, swing-hip joints track the desired footstep locations \cite{coros2010generalized, capturability2, faraji2014robust} and a task-space inverse kinematics method provides ground clearance \cite{faraji2014robust}. The entire method is currently based on position control, although it can be extended to torque control as well. The robot is always commanded to perform an in-place walking gait with the foot-placement and CoP damping \cite{sherikov2014whole, coros2010generalized} combined in both sagittal and lateral directions. These strategies bring stability against large momentary disturbances, and our walking gaits emerge from persistent sources of smaller disturbances, for example, external forces, titled torso angles and shifted pelvis positions. In these gaits, the foot-placement algorithm continuously interacts with the asymmetry sources. We call our method imprecise walking in the sense that we do not use the nominal 3LP walking gaits \cite{faraji20173lp} as feed-forward trajectories in our controller. The foot-placement controller is robust enough to handle violation of linearity assumptions, CoP damping, heel/toe motions and perturbations altogether. 

\subsection{Novelty}
Time-Projection control is the first novel aspect of this paper and gives an alternative to the state of the art MPC and capture-based controllers \cite{faraji2014robust,feng2016robust,capturability2}. The second novel aspect is that by applying time-projection on the 3LP model, we calculate simple control rules that account for swing dynamics and produce leg-retraction laws. These rules are very similar to capture-based laws \cite{capturability}, though with additional terms compensating for swing foot position and velocity errors. This second novel aspect provides a better dynamical match with the system (compared to the LIP model) and enables for an extremely robust bipedal walking behavior demonstrated on a simulated Atlas robot. We can produce gaits as fast as 1.7 m/s at 3 steps/s, and recover pushes up to 200 Ns strength. The self-selected speed in human is about 1.2 m/s at 1.9 steps/s for comparison \cite{bertram2005constrained}. Our simulations naturally produce dynamic walking with heel-contact, toe-off, vertical CoM, and torso motions altogether. We can also achieve step lengths longer than the leg length of the robot, indicating that our 3LP model and foot-placement rules do not impose themselves or their linearity assumptions on the robot and only suggest footstep locations. Finally, we also provide a software that calculates time-projection look-up-tables for arbitrary-sized robots, body-mass proportions and walking frequencies (generic for all walking speeds). With this flexibility, the swing dynamics of heavy-legged robots and fast stepping frequencies can be handled systematically. This paper starts with formulating the time-projection scheme and deriving foot-placement controllers in the next section. We continue by presenting walking and push-recovery results as well as a sensitivity analysis for the few involved parameters. Finally, we wrap up the paper by a discussion on performance and future directions to improve the proposed controller.

\section{Time-Projection}
\label{sec::theory}

To present the idea behind time-projection in a simple way, we consider a linear time-invariant system in which an error should be regulated to zero at certain sampling times. The 3LP model and hybrid phase-changes of walking are discussed later in this section, and the complex time-projection formulas of these systems are included in the appendix for further information.

\subsection{Linear System}
 Define a state vector $x(t) \in \mathbb{R}^{N}$ and a control vector $u(t) \in \mathbb{R}^{M}$. The system can be described by:
\begin{eqnarray}
\dot{x}(t) = a x(t) + b u(t)
\label{eqn::simple_system}
\end{eqnarray}
where $a \in \mathbb{R}^{N \times N}$ and $b \in \mathbb{R}^{N \times M}$ are constant matrices. The closed-form solution of this system at time $t$ is obtained by:
\begin{eqnarray}
x(t) = e^{at} x(0) + \int_0^t \! e^{a(t-\tau)} b u(\tau) \, \mathrm{d}\tau
\label{eqn::simple_system_solution}
\end{eqnarray}
For simplicity, we consider a constant input here, although this can be extended to linear or quadratic profiles without loss of generality. With a constant input, parametrized by the vector $U \in \mathbb{R}^{M}$, the equation (\ref{eqn::simple_system_solution}) takes the form:
\begin{eqnarray}
x(t) = e^{at} x(0) + (e^{at} - I)a^{-1}b \, U
\label{eqn::simple_system_constantU}
\end{eqnarray}
where we assumed $a$ is invertible. If $a$ is singular, a similar expression can be obtained by considering the Jordan form of $a$. We consider a period time $T>0$ at which the behavior of this system could be described discretely: 
\begin{eqnarray}
X[k+1] = A(T) X[k] + B(T) U[k]
\label{eqn::simple_system_discrete}
\end{eqnarray}
where $X[k]=x(0)$, $X[k+1]=x(T)$, $A(T) = e^{aT}$ and $B(T) = (e^{aT} - I)a^{-1}b$.

\subsection{DLQR Control}
Assume our linear system has a nominal solution $\bar{X}[k]$ and input $\bar{U}[k]$. Due to linearity, we can define error dynamics as follows:
\begin{eqnarray}
E[k+1] = A E[k] + B \Delta U[k]
\label{eqn::simple_system_error dynamics}
\end{eqnarray}
where $E[k] = X[k] - \bar{X}[k]$ and $\Delta U[k] = U[k] - \bar{U}[k]$. Here we dropped $(T)$ from $A(T)$ and $B(T)$ for simplicity. Assume this system is controllable and a DLQR controller can be found to minimize the following cost function and constraint:
\begin{eqnarray}
\nonumber &\underset{E[k],\Delta U[k]}{\text{min}} \sum_{k=0}^{\infty} E[k]^TQE[k]+\Delta U[k]^TR\Delta U[k] \\
& s.t. \, \begin{array}{l} E[k+1] = AE[k]+B\Delta U[k]  \end{array}
\, \, \,  k\ge0
\label{eqn::simple_system_lqr}
\end{eqnarray}
where $Q$ and $R$ are cost matrices. The optimal gain matrix calculated from this optimization is called $K \in \mathbb{R}^{M \times N}$, producing a correcting input $\Delta U[k] = -KE[k]$. This controller is only active at time instants $kT$ and provides corrective inputs to the system until the next sample $(k+1)T$. In this period, the effect of any disturbance on the system is cumulative, without any correction. Because of an exponential nature of (\ref{eqn::simple_system_solution}) in walking dynamics, even a small intermittent disturbance might create a substantial error at time $(k+1)T$. In non-hybrid systems, this issue might be fixed by increasing the resolution and redesigning the DLQR controller depending on disturbance dynamics. This is not possible for walking due to the hybrid nature. However, we can take advantage of a coarse-time DLQR controller and increase the resolution as described next.

\subsection{Time-Projection Idea}
\begin{figure}[]
	\centering
	\includegraphics[trim = 0mm 0mm 0mm 0mm, clip, width=0.5\textwidth]{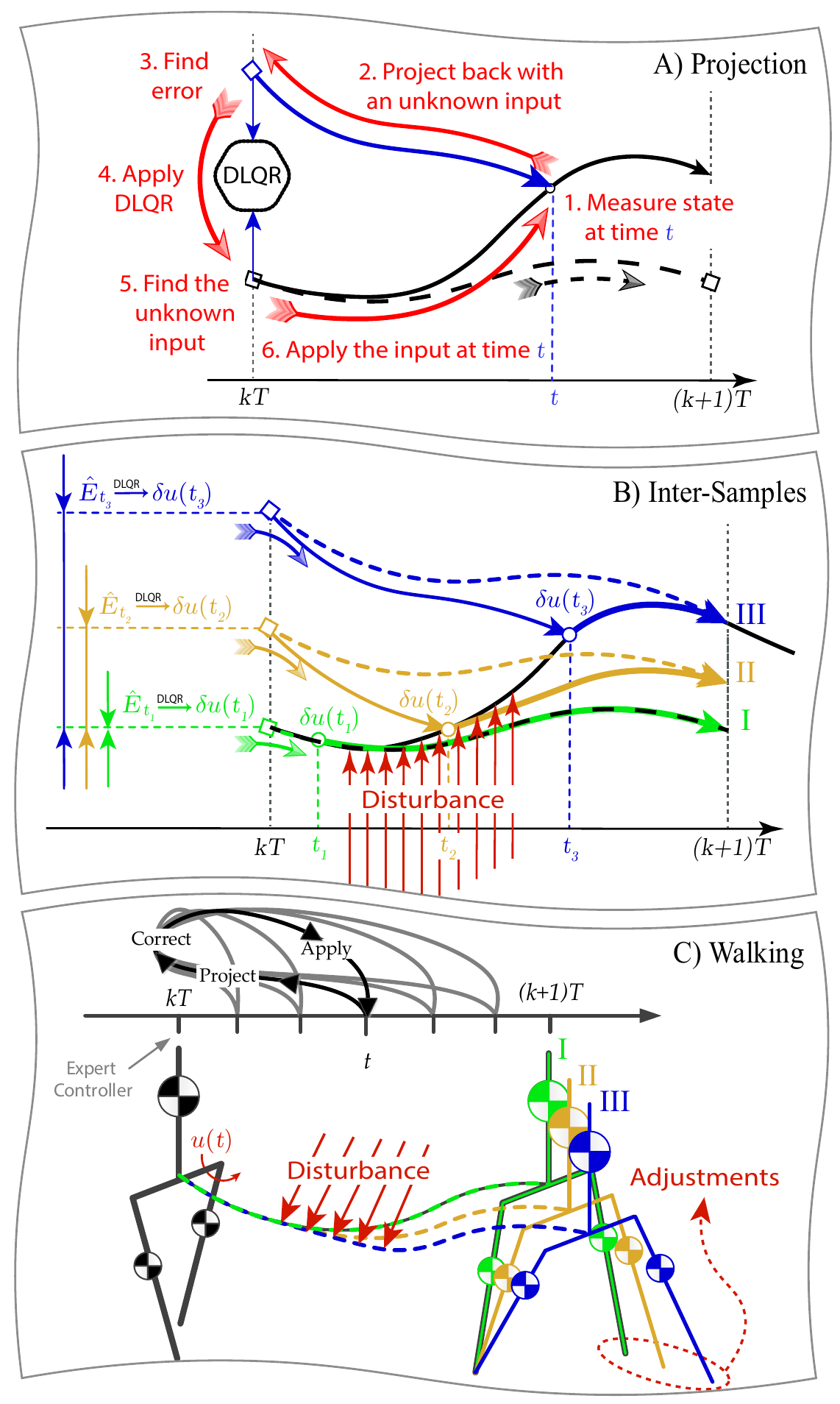}
	\caption{A) Schematic of time-projection during a single phase where the dashed and solid lines are nominal and actual trajectories. B) Without disturbance at $t_1$, time-projection produces no adjustment. During the disturbance, e.g. at $t_2$, this controller keeps updating the input. When the disturbance is finished, e.g. at $t_3$, the inputs remain constant until the next event. C) Trajectories of panel B) plotted in a walking system. Here the adjusted hip torques at times $t_1$, $t_2$ and $t_3$ result in different final footstep adjustments that can be given to a position-controlled robot.} 
	\label{fig::projecting_idea}
\end{figure}
Thanks to closed-form system equations (\ref{eqn::simple_system_constantU}), we can map the state measured at any time $t$ to the samples before and after. Fig. \ref{fig::projecting_idea}.A shows the time-projection idea between samples $kT$ and $(k+1)T$. In brief, to apply time-projection:
\begin{enumerate}
	\item Measure the current state $x(t)$ at time $t$.
	\item Project $x(t)$ back in time with an unknown $\delta \hat{U}_t[k]$ to find a possible initial state $\hat{X}_t[k]$.
	\begin{eqnarray}
	\nonumber x(t) =&& A(t-kT) \hat{X}_t[k] \\
	+ \, && B(t-kT) \, (\bar{U}[k]+\delta \hat{U}_t[k])
	\label{eqn::simple_system_xtoX+}
	\end{eqnarray}
	Here, the hat notation means that these variables are predicted at time $t$ and they are not actual system variables. The subscript $t$ indicates dependency on $t$.
	\item Now, given $\hat{X}_t[k]$ and $\bar{X}[k]$, find a projected error $\hat{E}_t[k]$ at the beginning of the phase. 
	\begin{eqnarray}
	\hat{E}_t[k] = \hat{X}_t[k] - \bar{X}[k]
	\label{eqn::simple_system_X-toE}
	\end{eqnarray}
	\item The expertise of DLQR controller can be used here to apply a feedback on $\hat{E}_t[k]$.
	\begin{eqnarray}
	\delta \hat{U}_t[k] = -K \hat{E}_t[k]
	\label{eqn::simple_system_dU}
	\end{eqnarray}
	\item Now, one can solve a system of linear equations to find $\delta \hat{U}_t[k]$. Defining $\bar{x}(t) = A(t-kT) \bar{X}[k] + B(t-kT) \bar{U}[k]$ and $e(t) = x(t) -\bar{x}(t)$ we have:
	\begin{eqnarray}
	\begin{bmatrix} A(t-kT) & B(t-kT) \\ K & I \end{bmatrix} \begin{bmatrix}
	\hat{E}_t[k] \\ \delta \hat{U}_t[k] \end{bmatrix} = \begin{bmatrix} e(t) \\ 0
	\end{bmatrix}
	\label{eqn::simple_system_solvedU}
	\end{eqnarray}
	\item The input is then directly transferred to time $t$ without any change and applied to the system:
	\begin{eqnarray}
	\delta u(t) = \delta \hat{U}_t[k]
	\label{eqn::simple_system_dUanswer}
	\end{eqnarray}
\end{enumerate}

This simple procedure can be done for any time instance $t$ between the coarse time-samples $kT$ and $(k+1)T$ to update the control input. The time-projection loop involves solving a linear system of equations whose dimensions depend on the number of control inputs and states in the system. Fig. \ref{fig::projecting_idea}.B demonstrates three different inter-sample times $t_1$, $t_2$ and $t_3$ before, in the middle and after the disturbance respectively. When following nominal trajectories, the time-projection controller produces zero adjustments $\delta u(t)$ indeed. During the disturbance, the controller keeps updating $\delta u(t)$ and after, it produces the same adjustment until the time $(k+1)T$.

\subsection{Single-Dimensional Stability}
\begin{figure}[]
	\centering
	\includegraphics[trim = 0mm 0mm 0mm 0mm, clip, width=0.5\textwidth]{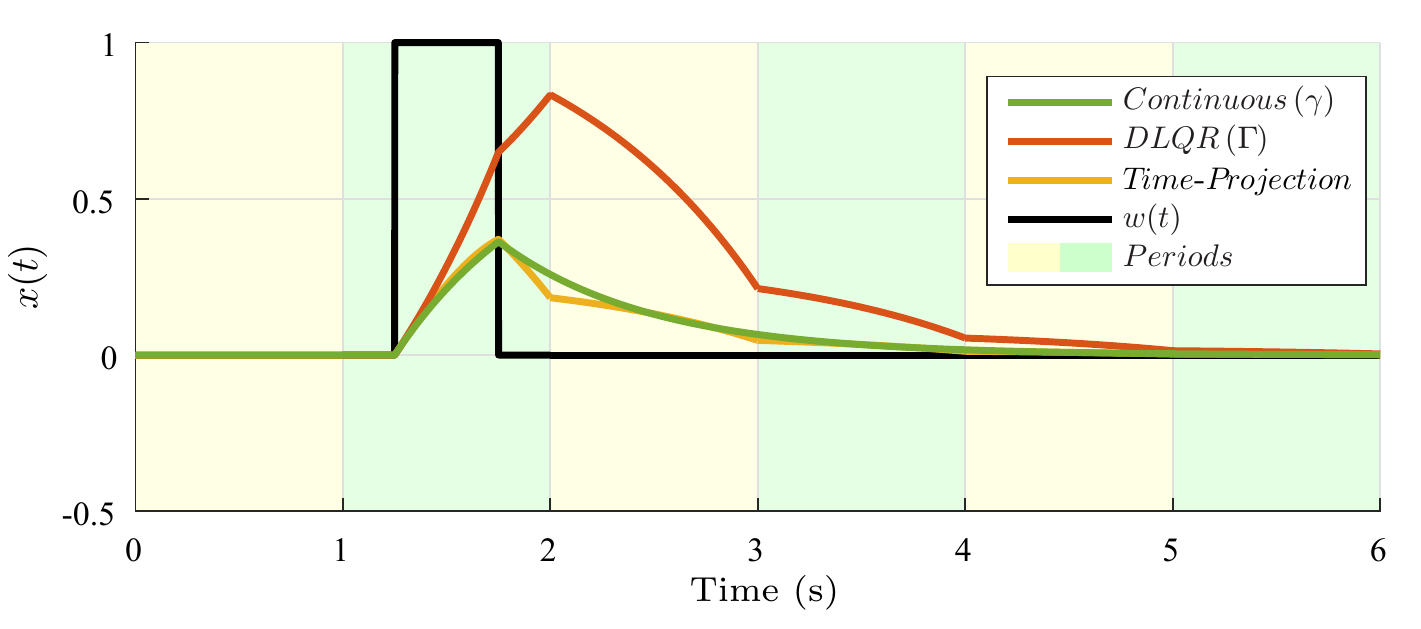}
	\caption{A one-dimensional system with continuous, DLQR and time-projection controllers. The DLQR controller overshoots due to a delayed response, though the time-projection controller performs very similar to a continuous control.} 
	\label{fig::1dof}
\end{figure}

To provide more intuition about closed-loop stability, we consider a simple system with a single state and a single input:
\begin{eqnarray}
\dot{x}(t) = x(t) + u(t) + w(t)
\label{eqn::1dof_system}
\end{eqnarray}
where $u(t)$ is the control input and $w(t)$ is the disturbance. Consider a control period $T$ and steady-state solutions $\bar{x}=0$ and $\bar{u}=0$. The closed-form evolution of this system could be written as:
\begin{eqnarray}
X[k+1] = e^TX[k]+(e^T-1)U[k]
\label{eqn::1dof_system_discrete}
\end{eqnarray}
assuming a constant $U[k]$ applied to the system. Now, imagine we use a discrete controller at time instances $kT$ to adjust the constant input with a law of:
\begin{eqnarray}
U[k] = -\Gamma X[k]
\label{eqn::1dof_system_lqr}
\end{eqnarray}
The stability of closed-loop system suggests that:
\begin{eqnarray}
|e^T- (e^T-1)\Gamma|<1
\label{eqn::1dof_system_lqr_stability}
\end{eqnarray}
which puts boundaries on $\Gamma$:
\begin{eqnarray}
1<\Gamma<\frac{e^T+1}{e^T-1}
\label{eqn::1dof_system_lqr_Gamma_bound}
\end{eqnarray}
For this system, the DLQR controller satisfies this criteria. The time-projection controller takes a state $x(t)$, maps it to the previous time-sample $kT$ and finds the control input based on the discrete controller $\Gamma$. Without loss of generality, we focus on the first sample ($k=0$) while other samples can be discussed by replacing $t$ with $t-kT$ in our equations. Time-projection infers:
\begin{eqnarray}
\nonumber x(t) &=& e^t \hat{X}_t[0] + (e^t-1)\delta \hat{U}_t[0]\\
\delta \hat{U}_t[0] &=& - \Gamma \hat{X}_t[0]
\label{eqn::1dof_system_projection}
\end{eqnarray}
where $\hat{X}_t[0]$ is a possible predicted initial state that can lead the current measurement $x(t)$. The two laws of (\ref{eqn::1dof_system_projection}) and system equations (\ref{eqn::1dof_system}) result in the following closed-loop system:
\begin{eqnarray}
\nonumber x(t) &=& e^t (-\frac{\delta \hat{U}_t[0]}{\Gamma}) + (e^t-1) \delta \hat{U}_t[0] \\
\dot{x}(t) &=& (1+\frac{1}{-\frac{e^t}{\Gamma} + e^t-1}) x(t)
\label{eqn::1dof_system_projection_closed_loop}
\end{eqnarray}
which is found by resolving $\hat{X}_t[0]$ in (\ref{eqn::1dof_system_projection}) and plugging $u(t) =\delta \hat{U}_t[0]$ in (\ref{eqn::1dof_system}). In order to produce finite feedbacks, the denominator in (\ref{eqn::1dof_system_projection_closed_loop}) should not have a zero in $0<t<T$. In other words, the root of denominator $t_0$ should be outside $[0,T]$:
\begin{eqnarray}
-\frac{e^{t_0}}{\Gamma} + e^{t_0}-1 = 0 \quad \rightarrow \quad t_0 = ln(\frac{1}{1-\frac{1}{\Gamma}})
\label{eqn::1dof_system_projection_root}
\end{eqnarray}
This leads to the following two conditions:
\begin{eqnarray}
\left\{ \begin{array}{l} ln(\frac{1}{1-\frac{1}{\Gamma}})<0 \rightarrow \frac{1}{1-\frac{1}{\Gamma}}<1 \rightarrow \Gamma<0 \\ ln(\frac{1}{1-\frac{1}{\Gamma}})>T  \rightarrow \frac{1}{1-\frac{1}{\Gamma}} > e^T \rightarrow \Gamma<\frac{e^T}{e^T-1} \end{array} \right. 
\label{eqn::1dof_system_projection_root_infinite}
\end{eqnarray}
which further tighten the boundaries of (\ref{eqn::1dof_system_lqr_Gamma_bound}) to:
\begin{eqnarray}
1<\Gamma<\frac{e^T}{e^T-1}
\label{eqn::1dof_system_lqr_Gamma_newbound}
\end{eqnarray}
Note that the DLQR controller does not necessarily satisfy this criterion, unless proper state and input cost matrices are chosen in the objective. Now, consider the Lyapunov function $V(t) = \frac{1}{2} x(t)^2$ whose derivative is:
\begin{eqnarray}
\dot{V}(t) = x(t) \dot{x}(t) = \epsilon (t) x(t)^2
\label{eqn::1dof_system_dlyap}
\end{eqnarray}
where:
\begin{eqnarray}
\epsilon (t) = 1+\frac{1}{-\frac{e^t}{\Gamma} + e^t-1}
\label{eqn::1dof_system_eps}
\end{eqnarray}
It is obvious that $\epsilon(t)$ monotonically decreases with time, because $\Gamma>1$. Thus:
\begin{eqnarray}
\dot{\epsilon} (t) = \frac{e^t(\frac{1}{\Gamma} - 1)}{(-\frac{e^t}{\Gamma} + e^t-1)^2} < 0
\label{eqn::1dof_system_deps}
\end{eqnarray}
Since $\epsilon(0) = 1-\Gamma < 0$, one can conclude that:
\begin{eqnarray}
\dot{V}(t) = \epsilon (t) x(t)^2 < \epsilon(0) x(t)^2
\label{eqn::1dof_system_lyapeps0}
\end{eqnarray}
which proves that $V(t)$ decreases over time. In the next phase, although $\epsilon (t)$ resets to its value at $t=0$ again, its derivative is always negative according to (\ref{eqn::1dof_system_deps}). Therefore, since $x(t)$ is continuous, the Lyapunov function $V(t)$ remains continuous at the phase transition moment and continues to decreases in the next phase. As a result, the system is stable with any choice of $\Gamma$ in (\ref{eqn::1dof_system_lqr_Gamma_newbound}). Note that in the absence of disturbances, for example between $t=2$ and $t=3$ in Fig. \ref{fig::1dof}, the time-projection controller always produces the same input, as if a DLQR controller was invoked at time $t=2$ and produced the same input. In these conditions, the DLQR controller behaves similarly and thus, it can be another guarantee of stability after $t=2$. In order to compare the DLQR and time-projection controllers with a simple continuous controller of $u(t) = -\gamma x(t)$, we find the continuous feedback gain $\gamma$ by converting closed-loop eigenvalues. The discrete eigenvalue ($\Lambda$) of (\ref{eqn::1dof_system_lqr_stability}) is converted to a continuous eigenvalue ($\lambda$) by a logarithm operation:
\begin{eqnarray}
\Lambda = e^T- (e^T-1)\Gamma = e^{\lambda T}
\label{eqn::1dof_system_discrete_eigen}
\end{eqnarray}
Given that $\lambda = 1 - \gamma$, we can find $\gamma$ as:
\begin{eqnarray}
\gamma = -ln(e^T-(e^T-1)\Gamma)/T+1
\label{eqn::1dof_system_continuous_eigen}
\end{eqnarray}
This let us compare the three types of controllers on a fair basis. Consider we select $T=1$, $Q=1$ and $R=1$ for the DLQR design which yields $\Gamma \simeq 1.43$ and satisfies (\ref{eqn::1dof_system_lqr_Gamma_newbound}). The equivalent continuous gain for this system would be $\gamma \simeq 2.37$. Fig. \ref{fig::1dof} demonstrates how an inter-sample disturbance could be rejected by these controllers. The DLQR controller overshoots, simply because of a late response which only starts at time $t=2$. The continuous controller $\gamma$ smoothly damps the disturbance while the time-projection controller performs damping in multiple steps almost with the same rate as the continuous controller. One can imagine that with decreasing $T$, the time-projection controller will converge to the continuous controller. 

\subsection{Multi-Dimensional Stability}
Stability analysis of multi-dimensional systems is very similar to our previous example. Assuming $k=0$ like before and zero reference trajectories, we can rewrite the equation (\ref{eqn::simple_system_solvedU}) with definitions of matrices $A$ and $B$:
\begin{eqnarray}
x(t) = e^{at} \hat{X}_t[0] + (e^{at} - I)a^{-1}b \hat{U}_t[0]
\label{eqn::multi_extended}
\end{eqnarray}
We multiply both sides by $-Ke^{-at}$ and obtain:
\begin{eqnarray}
-Ke^{-at}x(t) = -K \hat{X}_t[0] - K(I - e^{-at})a^{-1}b \hat{U}_t[0]
\label{eqn::multi_replaces}
\end{eqnarray}
Knowing from (\ref{eqn::simple_system_solvedU}) that $\hat{U}_t[0] = -K \hat{X}_t[0]$, we can write: 
\begin{eqnarray}
-Ke^{-at}x(t) = (I - K(I - e^{-at})a^{-1}b) \hat{U}_t[0]
\label{eqn::multi_final}
\end{eqnarray}
Which defines the matrix that should be inverted:
\begin{eqnarray}
M(t) = I - K(I - e^{-at})a^{-1}b 
\label{eqn::invertible matrix}
\end{eqnarray}
Finding eigenvalues of this matrix in closed-form as a function of time is impossible. However at time $t=0$, all these eigenvalues are equal to 1 which make the matrix invertible. At other times $0<t<T$ after calculating the DLQR gains $K$, we simply plot the eigenvalues numerically and verify that they do not pass through zero, otherwise we change the DLQR gains. Like the single-dimensional case, the time-projection controller produces constant inputs in the next phases in the absence of perturbations. These inputs are equal to the inputs that a DLQR controller would produce at the beginning and since the DLQR controller is stable, the time-projection controller is also stable. In the next section where application of time-projection in walking is presented, we introduce simple normalizations that leave a single parameter to tune in the DLQR cost function. For common humanoid sizes and human-like frequencies of walking, this parameter has a wide range of feasible numbers as will be shown in the results section.

\section{Walking Control}
\label{sec::control}

Up to this point, the concept of time-projection was only presented for continuous time-invariant systems. The hybrid nature of bipedal walking can be addressed via a simple transformation which hides model changes due to the hybrid switching modes. In this section, we apply the time-projection concept to our previously developed 3LP model for 3D walking \cite{faraji20173lp}. 3LP is composed of three linear pendulums with masses that remain in constant-height planes and inertias that approximate rotational dynamics. Thanks to linear properties, we can derive closed-form equations and obtain walking gaits at different velocities and frequencies. Consider we represent pelvis and feet positions by two-dimensional vectors $x_\text{Pelvis}(t)$, $x_\text{Swing}(t)$ and $x_\text{Stance}(t)$. Now define a 3LP state $x(t)$ by:
\begin{eqnarray}
x(t) = \begin{bmatrix} q(t) \\ \dot{q}(t) \end{bmatrix}, \quad
q(t) = \begin{bmatrix}
x_\text{Swing}(t)-x_\text{Stance}(t) \\ 
x_\text{Pelvis}(t) - x_\text{Stance}(t)
\end{bmatrix}
\label{eqn::3lp_state}
\end{eqnarray}
where $q(t) \in \mathbb{R}^{4}$. Now at phase change moments, a transformation $S$ can switch the legs:
\begin{eqnarray}
S = \begin{bmatrix} s & 0 \\ 0 & s \end{bmatrix}, \quad
s = \begin{bmatrix} -I_{2\times2} & 0 \\ -I_{2\times2} & I_{2\times2} \end{bmatrix}
\label{eqn::3lp_transform}
\end{eqnarray}
where $I_{2\times2}$ is the identity matrix and indeed $S^TS=I_{8\times8}$. With 3LP equations, we can find matrices $a(t)$ and $b(t)$ to describe swing leg and stance leg dynamics. The 3LP state $x(t)$ evolves in each phase according to (\ref{eqn::simple_system}) where $u(t)$ denotes swing hip torques. At the end of the phase, we can apply the matrix $S$ to the state vector and start the new phase with the same matrices $a(t)$ and $b(t)$. Therefore, although the 3LP model is a hybrid system, we can easily apply our previous theories on the system $Sa(t)$ and $Sb(t)$ which hides the hybrid modes. We refer to the original paper for extensive model details \cite{faraji20173lp} and only focus on compass walking and foot-placement strategies in this paper by disabling ankle torques in the 3LP model. Our Atlas controller, however, still benefits from some CoP damping as discussed later.

\subsection{Walking DLQR Design}
By defining a parametric piecewise linear profile for the hip torques, we can obtain a discrete system similar to (\ref{eqn::simple_system_discrete}) where $X[k] \in \mathbb{R}^{8}$ and $U[K] \in \mathbb{R}^{4}$ indicate initial state and input vectors in the phase $k$. These vectors include both sagittal and lateral directions. The matrices $SA(T)$ and $SB(T)$ are functions of step time $T$ and the robot's mechanical properties. There is also a constraint of $CX[k+1]=0$ (where $C \in \mathbb{R}^{2 \times 8}$) which imposes zero velocities for the swing foot at the end of the phase because there is no impact in our 3LP model. We postpone derivation of DLQR equations for constrained systems to Appendix \ref{sec::app_dlqr_const}. However, we note that the zero foot velocity constraint shapes our control laws depending on swing dynamics and DLQR cost function design. Our walking DLQR matrices $Q$ and $R$ normalize position errors by leg length $l$, velocity errors by $\sqrt{gl}$ and hip torques by $Mg$ where $g$ is gravity and $M$ is the total mass. These normalizations usually give very reasonable controllers, though we also multiply the matrix $R$ by a scalar $10^\mu$ to determine the importance of swing dynamics. The choice of $\mu=0$ normally works fine, but sometimes does not give an optimal stabilization as will be discussed in the results section.

\subsection{Walking Time-Projection}
\begin{figure*}[h!]
    \centering
    \includegraphics[trim = 0mm 0mm 0mm 0mm, clip, width=\textwidth]{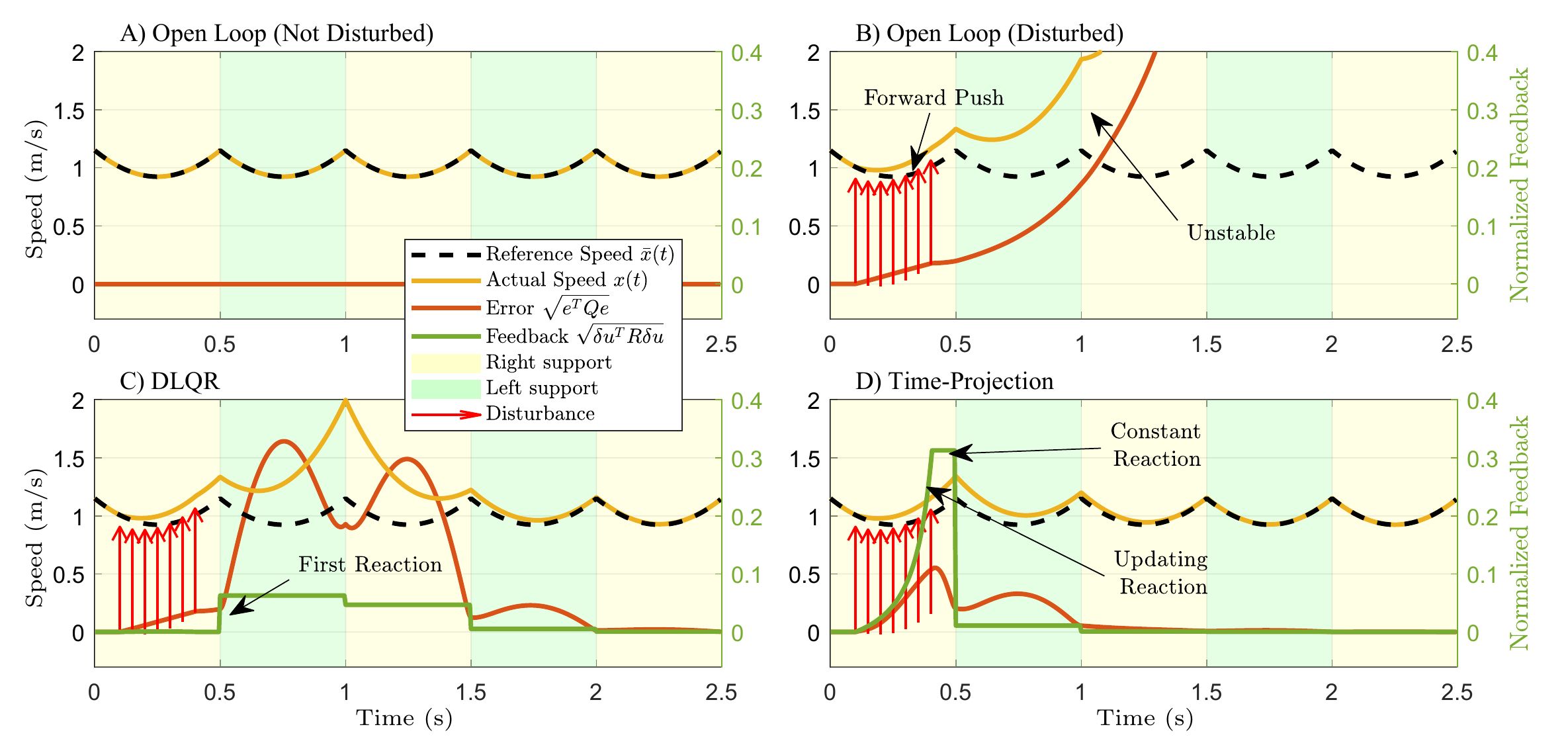}
    \caption{Trajectories of A) open-loop normal, B) open-loop disturbed, C) DLQR and D) time-projection controllers. Due to falling dynamics and unstable system modes, a moderate disturbance can lead to falling in few steps. The DLQR controller updates inputs only at phase-change moments which produces large deviations due to a delayed reaction. The time-projection controller, however, resolves this issue by a fast online reaction. Note that time-projection only translates continuous errors to discrete errors. The expertise of DLQR controller on discrete errors is then used to produce online corrections.} 
    \label{fig::projecting_walking}
\end{figure*}

\begin{figure*}[]
    \centering
    \includegraphics[trim = 0mm 0mm 0mm 0mm, clip, width=0.245\textwidth]{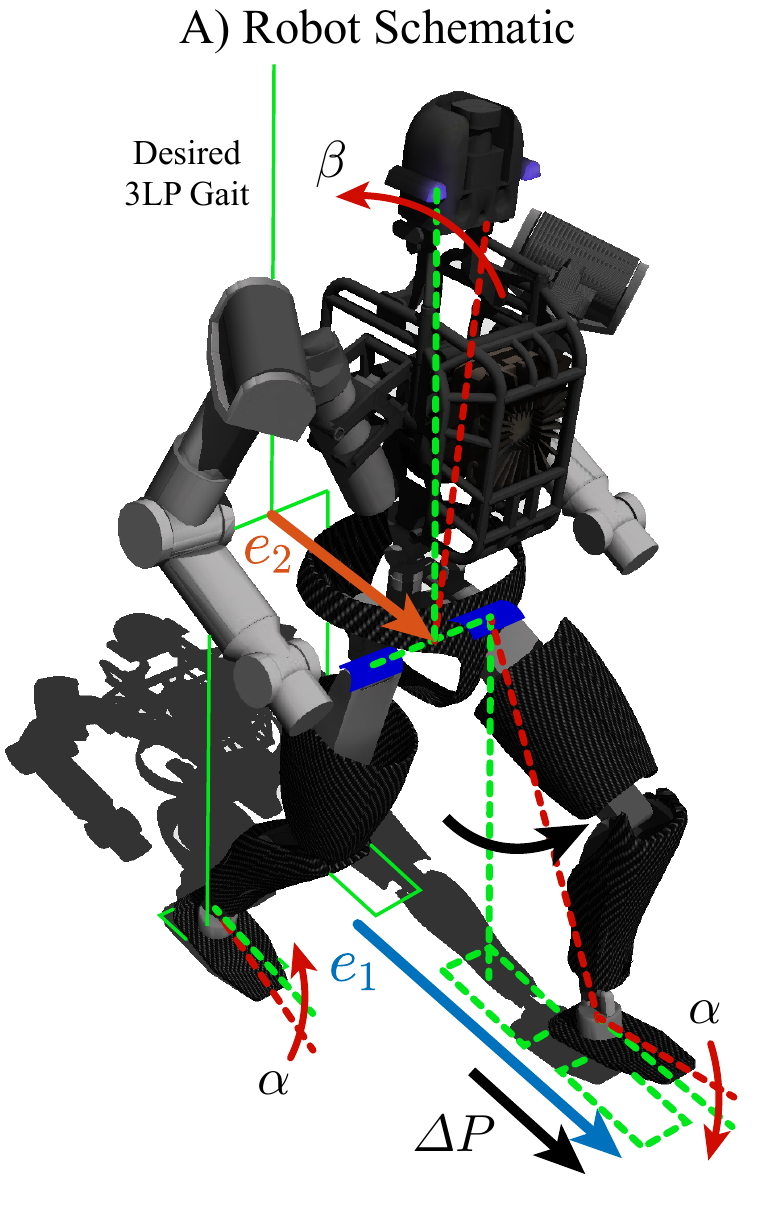} \includegraphics[trim = 0mm 0mm 65mm 0mm, clip, width=0.74\textwidth]{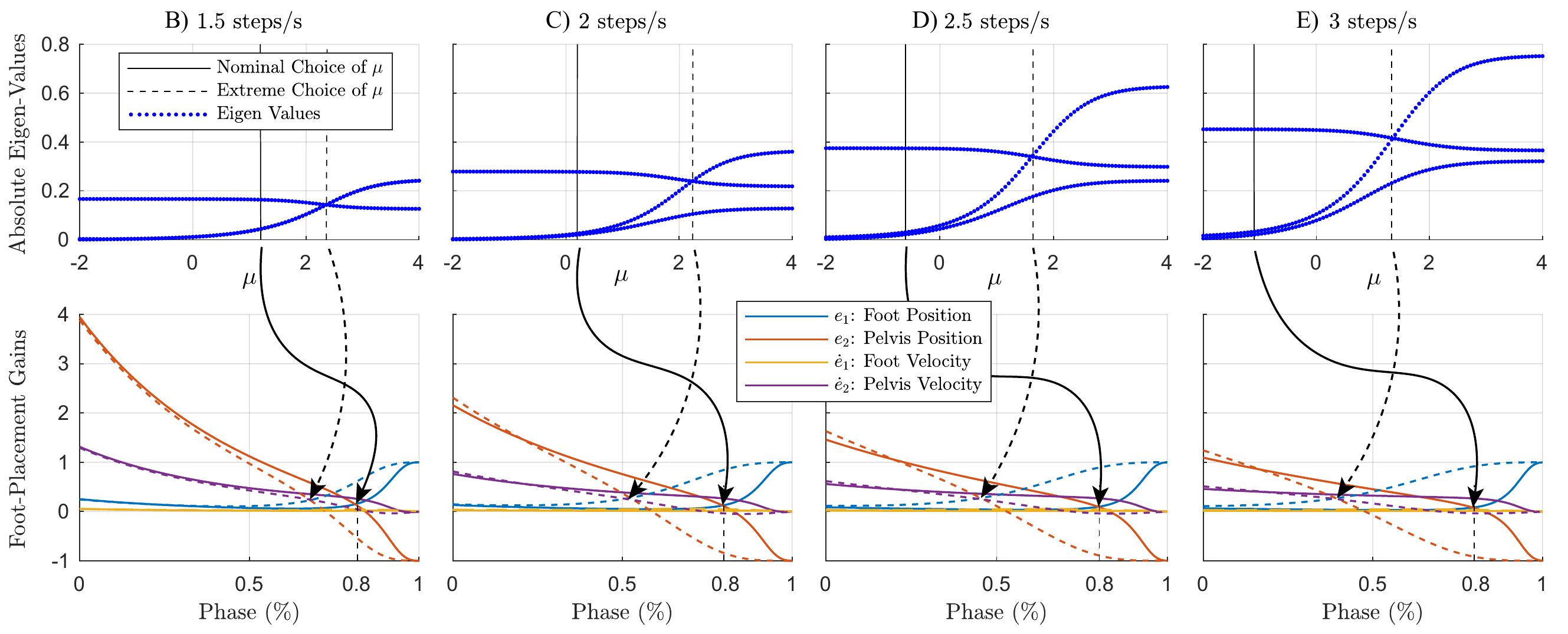}
    \caption{A) Atlas robot pushed from behind and falling forward while taking a new footstep. The desired gait here is shown in green. CoP damping gains $\alpha$ try to keep the feet flat while torso tracking gains $\beta$ keep the torso upright. B,C,D): Footstep adjustments $\Delta P(t)$ calculated by time-projection as a linear function of swing foot and pelvis deviations $e_1(t)$ and $e_2(t)$ and their derivatives. These curves are similar for both sagittal and lateral directions. At different walking frequencies, the parameter $\mu$ can tune the importance of swing dynamics and change the foot-placement curves. Our nominal choice of $\mu$ secures the last 20\% of the phase for swing foot stabilization. Higher or lower values can lead to a slower convergence (by larger eigenvalues) or ignorance of swing foot states.} 
    \label{fig::eigen}
\end{figure*}

At each instance of time during the phase, we can measure foot and pelvis deviation vectors $e_1(t) \in \mathbb{R}^2$ and $e_2(t) \in \mathbb{R}^2$ and their derivatives as shown in Fig. \ref{fig::eigen}.A. We can then apply time-projection at any time during the phase and obtain stabilizing hip torques as discussed earlier. The swing foot velocity constraints are also realized by simple manipulation of matrices as shown in Appendix \ref{sec::app_proj_const}. To better understand time-projection functionality, consider a 3LP gait with a frequency of 2 steps/s and a speed of 1 m/s. The open-loop velocities are shown in Fig. \ref{fig::projecting_walking}.A where the reference nominal trajectory $\bar{x}(t)$ matches the open-loop system behavior. Now, applying an inter-sample forward push in the first phase of motion results in an unstable deviation from nominal trajectories which eventually leads to falling as shown in Fig. \ref{fig::projecting_walking}.B. The DLQR controller can stabilize this system by adjusting the footstep at time $t=1$, shown in Fig. \ref{fig::projecting_walking}.C. This adjustment is a result of modifying the swing-hip torques at the beginning of the second phase (at $t=0.5$), where the accumulated deviation is first detected. As a result, corrective inputs $\Delta U$ are nonzero after the time $t=0.5$ until the system becomes stable completely. 

The time-projection controller has a different behavior, however. Without deviations, this controller produces no corrective input. Once a deviation appears, if there is no active disturbance applied to the system, the output of time-projection controller remains constant. However, once an external disturbance is present, the time-projection controller keeps updating the inputs which can be seen in Fig. \ref{fig::projecting_walking}.D. Notice the first phase where the external push starts acting on the system at $t\approx0.1$. At this time, corrective inputs start increasing until the disturbance disappears at $t\approx0.4$. After, the correction produced by the time-projection controller remains the same until $t=0.5$. In the next phase, the new stance foot location is already adjusted, and thus less effort is needed by the controller. Such footstep adjustment produces a larger error norm in the first phase compared to the DLQR controller, however, reduces this norm in the next phases considerably. As a result, the overall deviations from nominal trajectories are smaller in the time-projection controller, thanks to an online updating scheme.

To perform time-projection on position-controlled robots, we require desired swing-foot positions. Once we calculated instantaneous swing-hip torques like before, we can integrate the system forward to the end of the phase, and obtain final footstep adjustments. This process can be repeated for all the time-instances in the phase, leading to different final footstep adjustment values. Since the system is linear, these adjustments are linear functions of the initial error dimensions. Therefore, we can find time-variant coefficients and later apply them to arbitrary error values without needing to repeat the time-projection calculations online. The bottom row of Fig. \ref{fig::eigen} shows these coefficients or gains as a function of phase time. The errors $e_1(t)$ and $e_2(t)$ and their derivatives are multiplied by these gains and added together to form a step adjustment $\Delta P(t)$ shown in Fig. \ref{fig::eigen}.A. This figure only shows gains in the sagittal direction while lateral gains are exactly the same and applied on the second dimension of vectors $e_1(t)$ and $e_2(t)$. Finally, a truncation of $0.8l$ is applied to $\Delta P$ in both directions to avoid extra large step lengths and falling. Another truncation is also used on the lateral adjustments to prevent self-collision. It is worth mentioning that the curves in Fig. \ref{fig::eigen} are very similar if we scale the robot height for example to half of its size, and the walking frequency also proportionally according to our normalization coefficients.

\subsection{Leg Retraction}
Fig. \ref{fig::eigen} plots foot-placement gains at different walking frequencies and choices of the parameter $\mu$. On the top row of this figure, the largest closed-loop eigenvalue does not change much with $\mu$ whereas other eigenvalues start to increase with positive $\mu$ values. In these conditions, the parameter $\mu$ penalizes the hip torques exponentially and makes the swing leg move naturally. With smaller $\mu$ values, the zero foot velocity constraint is realized at the very end of the phase. Although the choice of $\mu=0$ normally gives very stable controllers, we slightly change it in each walking frequency to dedicate the last $20 \%$ of the phase for swing foot stabilization. In this case, the gain on $e_1$ converges to 1 while the gain on $e_2$ converges to -1 at the end of the phase. This property emerges from the zero foot velocity constraint in the 3LP model and makes the foot horizontally stationary (relative to the ground). It is called swing-leg retraction \cite{hobbelen2008swing} or ground-speed matching in the literature \cite{blum2010swing} and is known to increase gait stability. Due to missing swing dynamics, however, other template models like IP cannot produce this property mechanically and require intuitive rules in swing trajectory design \cite{hobbelen2008swing,faraji2014robust}. Our nominal choice of $\mu$ dedicates $80 \%$ of the phase to reach the target if tracking is slow, and $20 \%$ to stabilize in that position. Increasing the $20 \%$ phase reduces the convergence rate (eigenvalues start to increase in Fig. \ref{fig::eigen} top row). Decreasing the 20 \% stabilization phase also has the risk of ground-speed mismatch which may produce a horizontal impact, if not slipping.

\subsection{Swing Dynamics}
\begin{figure}[]
    \centering
    \includegraphics[trim = 0mm 0mm 0mm 0mm, clip, width=0.5\textwidth]{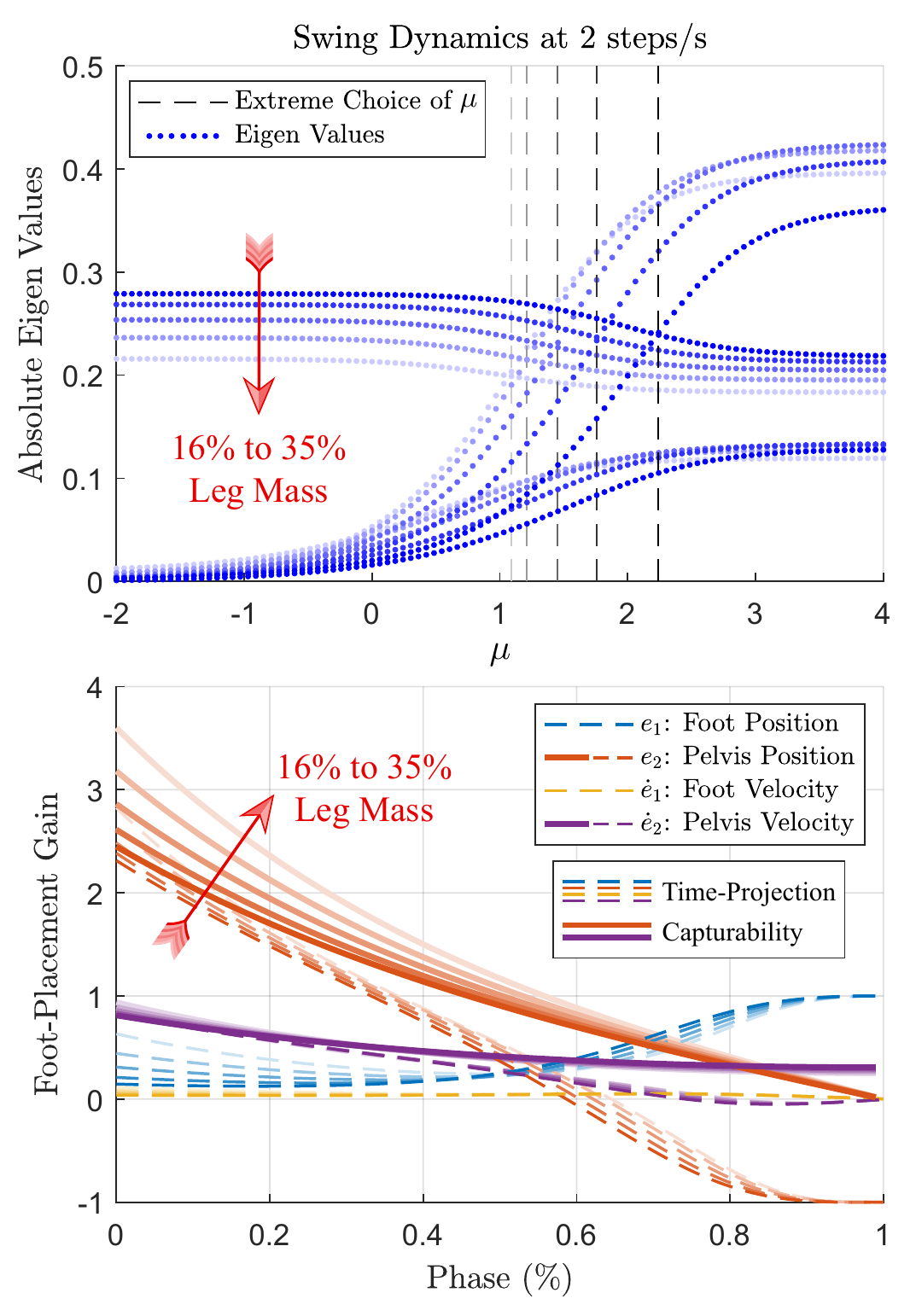}
    \caption{Foot-placement gains at 2 steps/s as a function of leg mass proportions. By moving the mass from the torso to the legs, we can artificially increase relative leg weights in the Atlas robot and recalculate time-projection gains. The capturability curves of the bottom plot indicate the effect of lowering the overall CoM height. Time-projection curves can, however, account for the increase in swing dynamics and provide optimum foot-placement laws.} 
    \label{fig::massprop}
\end{figure}

Fig. \ref{fig::eigen} also quantifies the effect of frequency on time-projection curves where the gains on the pelvis error get very large at lower frequencies. In these cases, an early-phase deviation requires larger reactions, since it grows exponentially in time due to unstable falling dynamics. Appendix \ref{sec::app_timing} provides a discussion on the sensitivity of inter-sample disturbance timings. Swing dynamics become important in both faster walking frequencies and heavy-legged robots at the same time \cite{faraji2018simple}. For this purpose, Fig. \ref{fig::massprop} quantifies time-projection gains for different body mass proportions. We also plot capturability gains \cite{capturability} on top to show which part of time-projection gains only come from CoM dynamics. In these plots, we artificially remove mass from the upper body and add it to the legs in the Atlas model. This brings the CoM lower and increases the gains on pelvis position errors. The robot geometry (leg length) is kept the same, however, to make the comparison fair. The choice of $\mu$ in Fig. \ref{fig::massprop} also corresponds to the point where two eigenvalues have equal magnitudes. While the foot stabilization part at the end of the phase does not change much in Fig. \ref{fig::massprop} bottom, we observe that the foot position gain increases with heavier legs, indicating a considerable swing dynamics. For example, when the swing foot is behind the stance foot, the sagittal part of $e_1$ becomes negative (refer to Fig. \ref{fig::eigen}.A) which produces a smaller adjustment $\Delta P(t)$. This means that the commanded final footstep position adapts to the current swing foot position and does not require a fast motion since swing dynamics is costly. In summary, derivation of time-projection gains on top of the 3LP model can optimize our controller for different robot dynamics.

\subsection{Robot Control}
In this paper, we use a simple low-level controller on a simulated Atlas robot to demonstrate the strengths of our straightforward foot-placement algorithm. We assume position control in the joints, availability of an IMU on the pelvis and contact force sensors in the feet. We only command in-place walking gait to the robot in all our scenarios with fixed timing. We use vertical sinusoids for foot trajectories in Cartesian space which are converted to joint-space by our position-based inverse kinematics library \cite{faraji2017singularity}.

\subsubsection{Torso Balancing Rules} 
Once we calculated the desired joint angles, we modify those in the hip joints according to the following rules. Denote the phase time by $0\le t<T$, the measured 3D contact force vectors by $F_{left}(t)$ and $F_{right}(t)$, actual IMU pitch angle by $\theta(t)$, desired pelvis angle by $\bar{\theta}(t)$ (the waist joints are disabled), and the required footstep adjustment by $\Delta P(t)$. We define a phase signal $\zeta(t)$ as:
\begin{eqnarray}
\zeta(t) = \frac{|F_\text{left}(t)|}{|F_\text{left}(t)|+|F_\text{right}(t)|}
\label{eqn::psi}
\end{eqnarray}
which then forms our modified left hip angles as:
\begin{eqnarray}
\begin{matrix}
\bar{q}(t) & = & (1-\zeta(t)) &  [\bar{q}_\text{IK}(t) - \theta(t) - \frac{\Delta P(t)}{l}] \\
 & + & \zeta(t) &  [q(t) + \beta(\theta(t)-\bar{\theta}(t))]
\end{matrix}
\label{eqn::hips}
\end{eqnarray}
Here, $q(t)$ is the measured and $\bar{q}_\text{IK} (t)$ is the desired left hip angle from inverse kinematics. A negative hip angle moves the leg forward according to our convention. For simplicity, equation (\ref{eqn::hips}) only mentions our strategy in the left sagittal hip joint while control rules for the other three hip joints are similar. Indeed for all other joints in the robot, $\bar{q}_\text{IK}$ is directly given to the joint PID controllers and not modified. The equation (\ref{eqn::hips}) tracks the desired pelvis angle $\bar{\theta}(t)$ with a gain $\beta$ in stance phase. Here, we feed the measured hip angle $q(t)$ back to the joint PID controller in order to only track the IMU angle and keep the pelvis (and torso) at $\bar{\theta}(t)$. In the swing phase, however, equation (\ref{eqn::hips}) moves the swing leg forward by compensating the actual rotation of the pelvis $\theta(t)$. Here $\Delta P(t)$ is divided by the leg length $l$ to approximate the desired hip attack angle. As mentioned earlier, the rhythm of motion is fixed in our controller, i.e. ground clearance motions and footstep adjustments are all synchronized with a desired step time of $T$. The signal $\zeta(t)$ is only used to smoothen unexpected phase transitions in perturbed walking conditions. In other words, a hip joint will not start balancing the torso until the foot in the same leg touches the ground.

\subsubsection{CoP Damping} 
In neutral conditions, the CoM in our robot will be on the line connecting the two ankles together. Flat feet and upright torso angles are part of the Cartesian tasks we give to inverse kinematics. With stiff position control, the ankle joints naturally resist perturbations thanks to having feet in the Atlas robot. During walking, however, this makes the feet always go on the edges which minimize the availability of stabilizing transversal wrenches. In this regard, we continuously add feedback on each desired foot orientation:
\begin{eqnarray}
\Delta_\text{ CoP}(t) = \begin{bmatrix} -\phi(t), \ - \alpha \theta(t),\ 0 \end{bmatrix}
\label{eqn::thetaadaptive}
\end{eqnarray}
where ${\phi}(t)$ and ${\theta}(t)$ are measured roll and pitch angles of that foot after state estimation (registered in the world-frame by IMU angles). The equation (\ref{eqn::thetaadaptive}) regulates roll and patch angles with gains of 1 and $\alpha$ which is also normally set to 1. This equation removes ankle joint stiffness and simply introduces a damped behavior. It makes the feet always approaching horizontal orientations with certain dynamics, depending on the joint's internal damping which is set to 0.1 Nms/rad in our simulations. Later in the results section, we will show how changing the parameter $\alpha$ would influence the walking style, i.e., heel-contact and toe-off phases. This gives better insight into how this parameter should be tuned on a real robot.

\subsubsection{State Estimation} 
This block in our controller is as simple as calculating forward kinematics with joint encoders and IMU measurements. To apply time-projection control, we only calculate relative horizontal pelvis and foot positions and apply time-differentiation to calculate their derivatives. Earlier in \cite{faraji2015practical}, we developed an advanced Kalman filter which calculated the leg kinematic chains from CoP points, allowing the foot to tilt on the edges during locomotion. However, in this work, we only calculate forward kinematics for the point under the ankle joint axis. During heel-contact and toe-off phases, this point may not be on the ground, but it gives a reasonable approximation of the support location. Our foot-placement algorithm, therefore, does not consider the actual CoP as the support point like \cite{sugihara2017foot} and ignores the small stabilization coming from the CoP damping rules.

\subsubsection{Heading Stabilization} 
Since heel-contact and toe-off phases appear during locomotion, the robot might slightly slip and turn on the feet edges. To correct for such deviations, we add feedback on the desired Cartesian foot orientations. For this purpose, denote the desired pelvis heading by $\bar{\psi}(t)$, the actual IMU heading by $\psi(t)$ and a correction term by:
\begin{eqnarray}
\Delta_\text{ heading}(t) = \begin{bmatrix} 0, \ 0,\ \eta(\bar{\psi}(t)-\psi(t)) \end{bmatrix}
\label{eqn::headingcorrect}
\end{eqnarray}
where the parameter $\eta=0.3$ brings enough stability in all our walking simulations. Desired tasks are then defined as:
\begin{eqnarray}
\nonumber \bar{R}_\text{ left}(t) &=& \Delta_\text{ CoP, left}(t) + (1-\zeta(t))\Delta_\text{ heading}(t) \\
\nonumber \bar{R}_{ right}(t) &=& \Delta_\text{ CoP, right}(t) + \zeta(t)\Delta_\text{ heading}(t) \\
\bar{R}_\text{ pelvis}(t) &=& \Delta_\text{ heading}(t)
\label{eqn::heading}
\end{eqnarray}
where $\bar{R}(t)$ denotes desired Cartesian orientations given to our inverse kinematics. These equations apply corrections only when the feet are in contact with the ground. The signal $\bar{\psi}(t)$ can indeed change with time, though our controller treats the turning tasks as unwanted perturbations, not optimally planned footstep locations like our previous work with MPC \cite{faraji2014robust}. In the next section, we will demonstrate a turning scenario together with our walking and push-recovery simulations.

\section{Results}
\label{sec::results}

After calculation of in-place walking trajectories in the Cartesian space, state estimation and application of CoP damping and heading stabilization rules, we convert the resulting Cartesian tasks to joint-space via inverse kinematics and modify the desired hip joint angles according to (\ref{eqn::hips}). These angles are then given to the position controllers of Atlas which run at 100 Hz in our simulations. The robot therefore only walks in-place unless some intermittent or persistent source of perturbation make it move. The nominal choices of controller parameters are mentioned in Table \ref{tab::param}. We will demonstrate walking and push-recovery scenarios in this section as well as sensitivity studies for parameters $\alpha$, $\beta$ and $\mu$ which are ankle and hip gains, and the DLQR cost design parameter. We also generate gaits at different frequencies, compare our controller with the capture-based method and demonstrate turning gaits. Extensive videos of these scenarios can be found in the accompanying video.

\begin{table}
    \begin{center}
        \caption{Nominal choices of controller parameters.}
        \label{tab::param}
        \begin{tabular}{ccc}
            Parameter & Function & Nominal Choice \\
            \hline 
            $1/T$ & Stepping frequency & 2 steps/s \\
            $\alpha$ & Ankle damping & 1 \\
            $\beta$ & Torso tracking gain & 2 \\
            $\mu$ & Swing dynamics cost & 20\% stabilization \\
            $\eta$ & Heading stabilization & 0.3 \\
            \hline
        \end{tabular}
    \end{center}
\end{table}

\subsection{Push-Recovery}
\begin{figure}
    \centering
    \includegraphics[trim = 0mm 0mm 0mm 0mm, clip, width=0.5\textwidth]{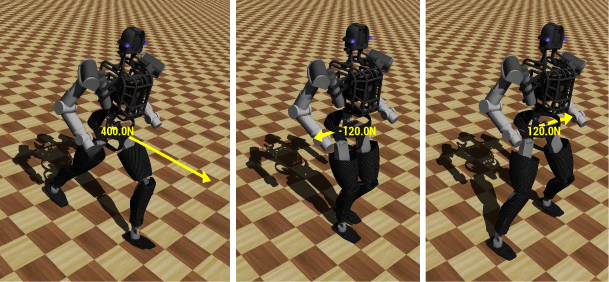}
    \caption{The first step taken by the left leg in our push-recovery scenarios. Here, forces are applied for a duration of $T=0.5$ s during the phase. The proposed method is able to recover impulses up to 200 Ns in the sagittal and 60 Ns in the lateral directions.} 
    \label{fig::pushrecovery}
\end{figure}

In our Atlas controller, both the CoP damping and hip control laws allow the robot to fall freely in different directions. However, depending on the measured deviation from in-place walking which is shown in Fig. \ref{fig::eigen}.A, the time-projection algorithm adjusts foot locations and tries to capture the motion. Fig. \ref{fig::pushrecovery} shows examples of large external pushes applied to the pelvis for a duration of $T=0.5$ s. The largest recoverable impulse in our method is about 200 Ns in forward and backward directions while the MPC-based method of \cite{feng2016robust} for example recovers 72 Ns impulses at maximum. The maximum recoverable lateral impulse in our method is also about 60 Ns while the variable-time method of \cite{khadiv2016step} can recover 20 Ns on a simulated Sarcos humanoid. Maximum lateral impulses are more limited due to self-collision constraints in case of opposite pushes, smaller CoP damping, and the fact that the CoM is normally between the two feet and already falling in the direction of non-opposite pushes. The recovery process may take longer when a self-collision is prevented, or during a backward push where the CoP damping cannot help much due to a small heel size. A short analysis of viable states is provided in Appendix \ref{sec::app_viability} to clarify the strength of foot-placement alone which has the dominant role here compared to the CoP damping strategy. We refer to the accompanying video for demonstrations of push-recovery scenarios in all four directions. 

\subsection{Walking}
\begin{figure*}
    \centering
    \includegraphics[trim = 0mm 0mm 0mm 0mm, clip, width=1\textwidth]{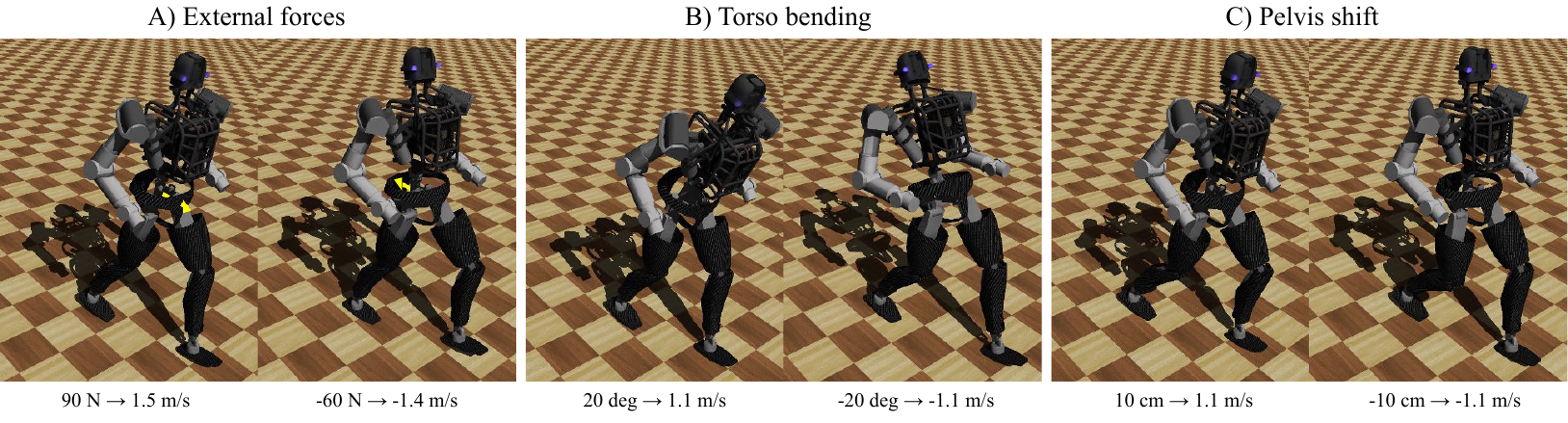}
    \caption{Walking gaits emerged from persistent sources of asymmetric disturbances: A) external pushes, B) torso tilts and C) pelvis shifts. In these cases, we do not command a desired walking gait and let it emerge automatically.} 
    \label{fig::walking}
\end{figure*}

Walking gaits can be produced by persistent sources of asymmetry or perturbation in the system. Here, we do not command the desired velocity. Instead, we apply persistent external pushes, or tilt the torso, or shift the pelvis, i.e., moving it forward or backward before inverse kinematics. The neutral pelvis position in our in-place walking gait brings the CoM on top of the ankle joints while a shift by few centimeters can make the robot fall and thus trigger a walking gait. In these scenarios, walking emerges through continuous interactions between the perturbation and the foot-placement controller. The CoP damping rules also contribute slightly. Fig. \ref{fig::walking} shows example snapshots in all these three cases while the controller remains the same. The robot always resumes to in-place walking when the asymmetry source disappears. Indeed, similar strategies like moving the arms forward or adding a manual offset to the ankle joints can lead to similar walking gaits. We refer to the accompanying video for full demonstrations of these walking gaits. Recovering from intermittent disturbances during such persistent asymmetries is also possible. For example, we can produce walking by shifting the pelvis forward and apply short external forces during the gait. A video of push-recovery during walking is included in the accompanying video.

\subsection{CoP Damping Sensitivity}
\begin{figure}
    \centering
    \includegraphics[trim = 0mm 0mm 0mm 0mm, clip, width=0.48\textwidth]{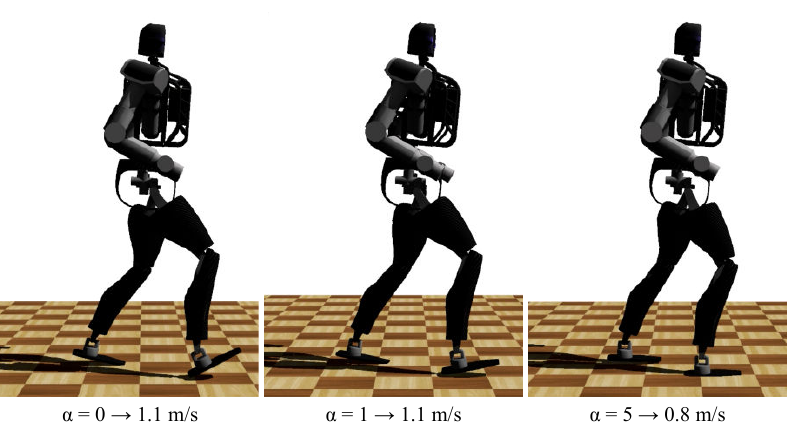}
    \caption{The effect of parameter $\alpha$ on gait geometry. Here the walking gait is produced by the pelvis shift strategy. The choice of $\alpha=0$ freezes the stance ankle while $\alpha=5$ makes the foot quickly adapt to the ground.} 
    \label{fig::anklegain}
\end{figure}

Remember that parameter $\alpha$ determines the amount of CoP damping in the ankles. With $\alpha=0$, the feet do not adapt and always go on the edges. With a large choice of $\alpha=5$ on the other hand, the feet always become flat and less human-like. All these scenarios are shown in Fig. \ref{fig::anklegain} where the walking gait is produced by shifting the pelvis. Flat feet conditions naturally produce higher vertical bounces in the pelvis, because the stance leg rotates around the ankle all the time and cannot go on the toes during phase transitions. Such pronounced vertical bounces lead to larger touch-down impacts in our controller which capture some energy and resist against forward progression. The choice of $\alpha=0$ is more unstable due to transversal slippages which cannot be completely compensated because our heading stabilization also requires short phases of complete foot contact with the ground. Videos of these gaits are included in the accompanying video.

\subsection{Torso Balancing Stiffness}
\begin{figure}
    \centering
    \includegraphics[trim = 0mm 0mm 0mm 0mm, clip, width=0.48\textwidth]{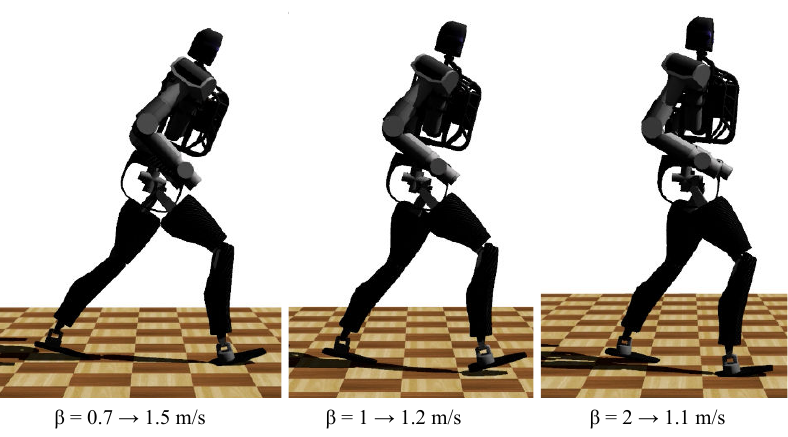}
    \caption{The effect of parameter $\beta$ on gait geometry. Reducing this parameters in the stance-hip controller can make the torso free which then bends forward or backward in the direction of motion. } 
    \label{fig::hipgain}
\end{figure}

The stance-hip in our controller tracks desired torso angles by the gain $\beta$ which is normally set to 2 in our controller. During walking gaits produced by the pelvis shift strategy, interestingly, lowering the gain $\beta$ can lead to torso bending in the direction of walking, which is a feature of human walking too \cite{geyer2010muscle}. In our controller, this phenomena further increase the walking speed as observed in Fig. \ref{fig::hipgain}, e.g. in case of $\beta=0.7$. The foot-placement controller is yet able to stabilize the motion. However, reducing $\beta$ would make the torso oscillate and perturb the gait considerably. The choice of $\beta=2$ minimizes torso bending while values higher than this point do not track desired torso angles any better and often make the controller unstable. The accompanying video contains the scenarios of Fig. \ref{fig::hipgain} together with a larger choice of $\beta=3$ in which the robot stumbles during walking.

\subsection{DLQR Cost Design}
\begin{figure}
    \centering
    \includegraphics[trim = 0mm 0mm 0mm 0mm, clip, width=0.5\textwidth]{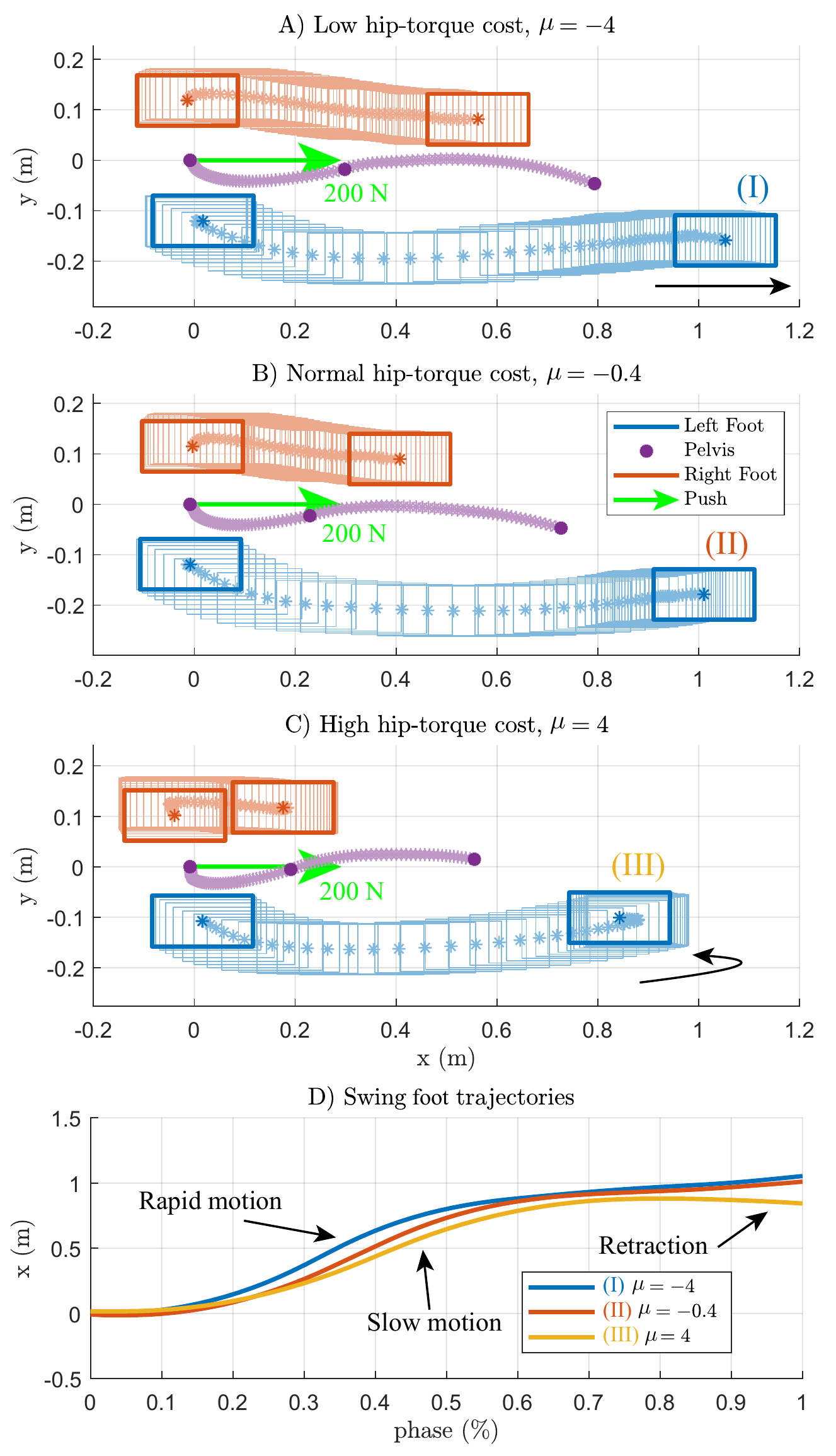}
    \caption{The effect of parameter $\mu$ on foot-placement behavior. A small $\mu$ in A) leads to fast reactions while a large $\mu$ in C) makes the swing foot very slow with a considerable leg-retraction. Our nominal choice in B) provides enough swing speed and 20\% final stabilization. D) The sagittal right foot position in these scenarios.} 
    \label{fig::DLQRgain}
\end{figure}

Remember from Fig. \ref{fig::eigen} that the parameter $\mu$ can penalize hip torques and determine the importance of swing dynamics. To better show this effect, we applied a push of 200 N to the robot for $T=0.5$ s during in-place walking (at 2 steps/s) and plotted the first two footsteps for different choices of $\mu$. Fig. \ref{fig::DLQRgain}.B shows our nominal choice of $\mu=-0.4$ which provides 20\% final stabilization and moderately captures the push. A smaller choice of $\mu=-4$ in Fig. \ref{fig::DLQRgain}.A let the robot take a larger first step first and then a similar second step. However, a large choice of $\mu=4$ in Fig. \ref{fig::DLQRgain}.C moves the first step only slightly forward while the second step follows its natural dynamics. This large choice of $\mu$ always leads to leg-retraction, often moving the leg backward by penalizing the swing-hip torques considerably like a free pendulum. Fig. \ref{fig::DLQRgain}.D compares these three scenarios by plotting the sagittal position of the right foot in time. Our nominal choice of $\mu=-0.4$ provides reasonable swing speeds and enough ground-speed matching or leg-retraction. Videos of these scenarios are also included in the accompanying video.

\subsection{Stepping Frequency}
\begin{figure}
    \centering
    \includegraphics[trim = 0mm 0mm 0mm 0mm, clip, width=0.5\textwidth]{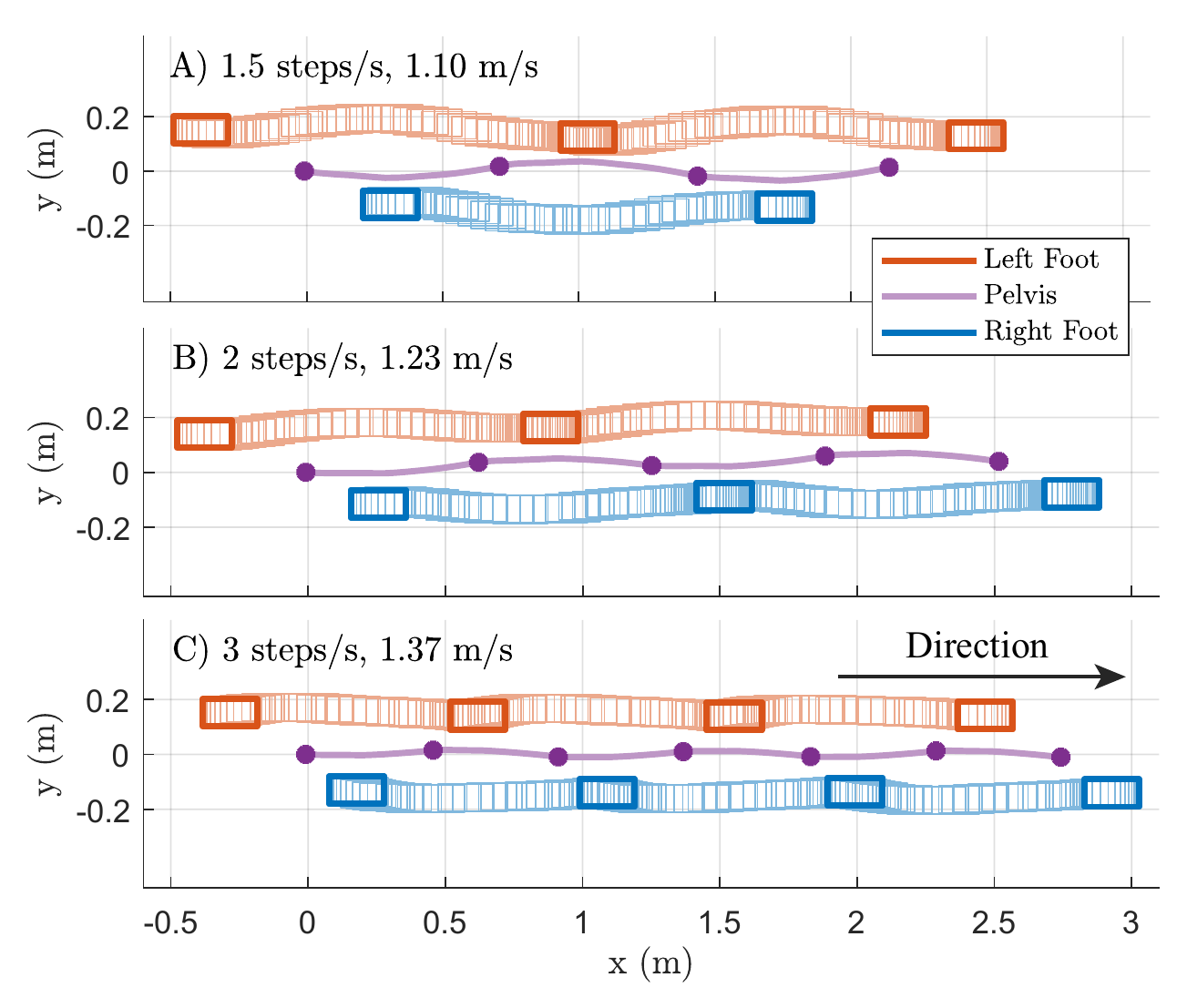}
    \caption{Walking gaits produced by the pelvis shift strategy at different stepping frequencies. At a frequency of 1.5 steps/s in A), the step length is about 60\% more than the leg length which indicates violation of constant-CoM assumptions in the 3LP model. The faster frequency of 3 steps/s in C) requires shorter steps though and less lateral bounces, producing smoother motions in the limbs.} 
    \label{fig::freq}
\end{figure}

Our controller can modulate the stepping frequency and reach slower or faster rhythms of motion. Fig. \ref{fig::freq} shows example gaits produced by the pelvis shift strategy at about the same speed. In these cases, we have changed the parameter $\mu$ for each case according to Fig. \ref{fig::eigen} to make the controller more optimal, although walking with $\mu=0$ is also possible in all of these frequencies. Videos of these scenarios are included in the multimedia attachment. A frequency of 3 steps/s in Fig. \ref{fig::freq}.C is very close to human's maximum frequency \cite{bertram2005constrained}. Our maximum speed at this frequency is about 1.7 m/s (shown in Fig. \ref{fig::capture}) while human can reach up to 2.34 m/s or even faster \cite{bertram2005constrained}. The robot easily goes on the toes which let it take longer steps during dynamic walking. Our CoP damping strategy is too simple to apply push-off forces and produce more human-like shank-foot coordinations \cite{cappellini2006motor}. We consider adding push-off strategies in future work to reduce vertical pelvis motions geometrically and enable for even longer step lengths. On the lower end also, we cannot go below 1.5 steps/s whereas the human can perform walking at 0.8 steps/s or even slower \cite{bertram2005constrained}. At these frequencies again, our CoP control strategy is too simple to provide fast balancing reactions unless we increase the damping. A transition to stand-still can also be realized by switching off the CoP damping completely.

\subsection{Comparison with Capturability}
\begin{figure}
    \centering
    \includegraphics[trim = 0mm 0mm 0mm 0mm, clip, width=0.5\textwidth]{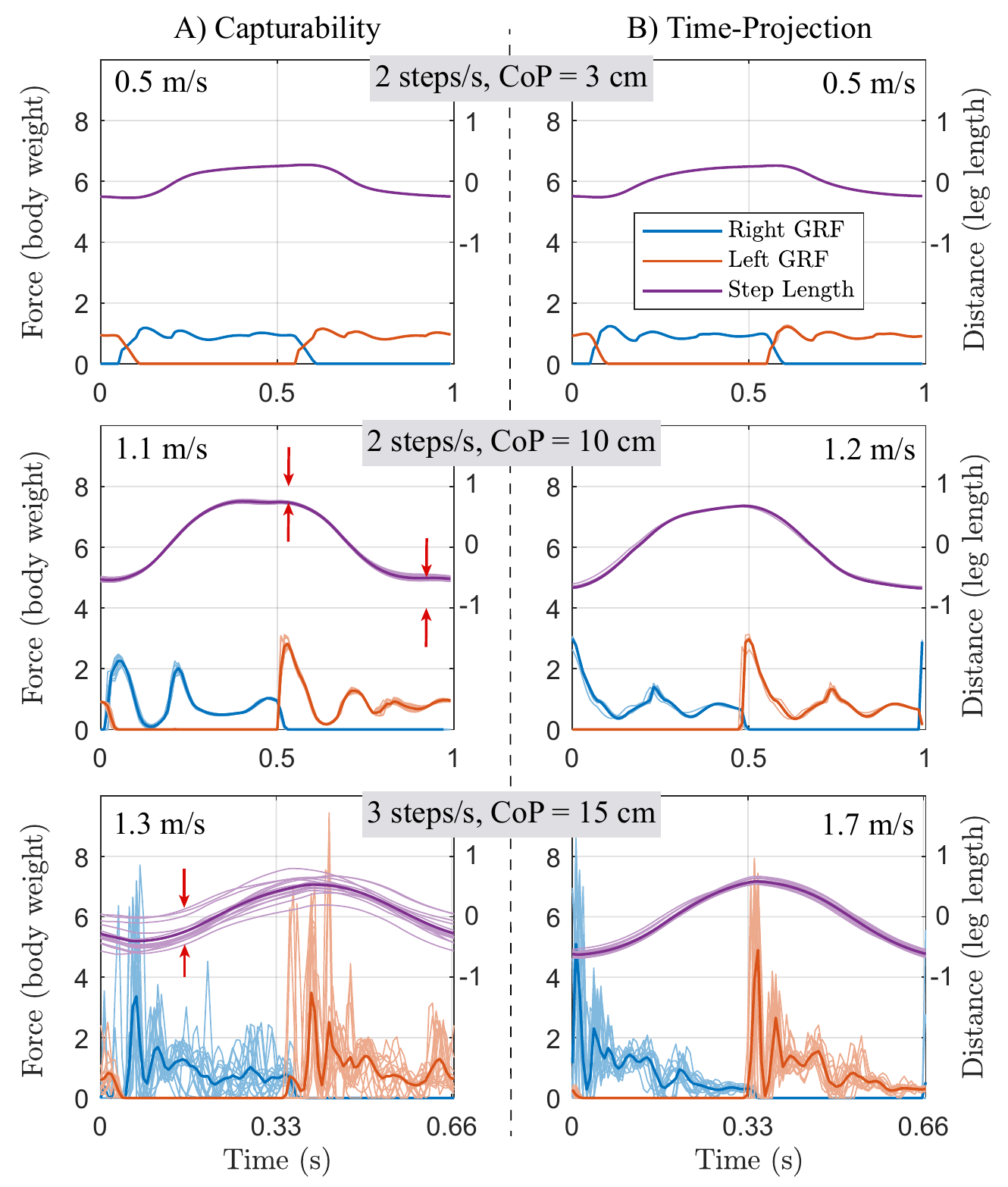}
    \caption{Comparison of capturability and time-projection methods for foot-placement. The plots here present at least 15s of walking with the pelvis shift strategy. The capturability method performs very similar to time-projection in slow walking gaits shown on top. However, it converges to an asymmetric gait in faster speeds shown in the middle. It also produces large gait variations at 3 steps/s where swing dynamics becomes more important.} 
    \label{fig::capture}
\end{figure}

Keeping the low-level controller the same, in this part, we attempt to apply capturability rules instead of time-projection. The goal is to find boundaries of swing dynamics, since the capturability gains of Fig. \ref{fig::massprop} on the LIP model do not account for swing foot errors and combine them with CoM dynamics. Fig. \ref{fig::capture} shows walking scenarios produced by the pelvis shift strategy at different stepping frequencies. In slow walking gaits, the two controllers perform almost similar with no variations. By increasing the pelvis shift further in the middle row of Fig. \ref{fig::capture}, the time-projection controller can still produce walking. However, the capturability method converges to a variable and asymmetric gait quantified by different left/right step lengths. Eventually, by increasing the frequency to 3 steps/s which requires considerable swing dynamics \cite{faraji2018simple}, we observe that the capturability method produces considerable gait variations due to the systematic absence of swing dynamics. Videos of these walking gaits are included in the accompanying video. 

\subsection{Turning}
\begin{figure}
    \centering
    \includegraphics[trim = 0mm 0mm 0mm 0mm, clip, width=0.5\textwidth]{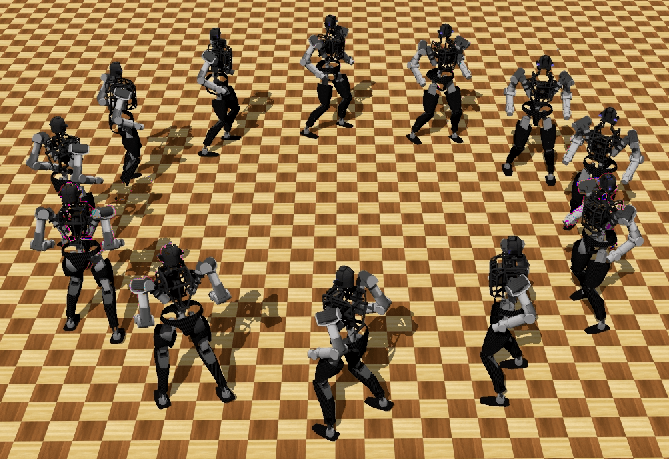}
    \caption{Snapshots of a forward walking and turning gait emerged from the pelvis shift strategy and commanding a desired heading signal.} 
    \label{fig::turn}
\end{figure}

As mentioned in the previous section, our heading stabilizer can track moderate turning commands by treating this signal as a perturbation. Since the original 3LP model has a pelvis segment, it cannot turn due to the nonlinearities involved. However, foot-placement rules derived from the 3LP model are generic and can stabilize the robot in both directions. With this feature, we can command the robot to turn with a moderately changing heading angle and expect the foot-placement laws to stabilize the gait. Example snapshots are demonstrated in Fig. \ref{fig::turn} with videos included in the accompanying video.

\section{Discussions}
\label{sec::conclusion}

The theory we presented in this paper forms a robust walking controller with a large viable set of states. This controller is based on foot-placement, a walking control strategy which only works based on the hybrid nature of walking and requires future predictions. We developed time-projection as an alternative to MPC to account for this requirement and embedded all the knowledge of future predictions in the DLQR controller. Time-projection is, therefore, a simple translation of continuous errors to the discrete space where the expertise of DLQR can be utilized. We also applied time-projection to the 3LP model which adds swing and torso balancing dynamics to the original LIP model. Our 3LP model produces footstep adjustments that naturally consider swing dynamics, hip torques costs and leg-retractions. 

\subsection{Imprecise Walking}
Using a very simple low-level Atlas controller, we admit that we do not precisely realize 3LP trajectories on the full robot. In other words, we do not accurately track any desired 3LP posture, or we do not keep the CoM height constant. Our Atlas controller works very similarly to an ordinary inverted pendulum with fixed leg lengths. We allow the pelvis to go up and down, the feet to go on the edges and the swing leg to touch the ground with perturbed timing. All these features make the robot more human-like, though violate the assumptions behind 3LP model. Our walking gaits also emerge from continuous interactions between disturbances and foot-placement. Therefore, our powerful controller overshadows such assumption violations. For this reason, we call our method "imprecise walking". A similar concept is also used in \cite{sugihara2017foot, sherikov2014whole, capturability2} where a stop condition is always forced as a terminal constraint in a number of future steps to capture the motion. In these methods, other user-defined tasks \cite{sherikov2014whole} or tunable gains \cite{capturability2} let the robot walk and move forward before reaching the terminal constraint.

\subsection{CoP damping}
A large body of humanoid walking literature is based on CoP control and the use of ankle torques. We combined this strategy with our foot-placements through simple CoP damping rules. The behavior of ankle joints in our robot is tunable and allows for heel-contact and toe-off phases \cite{sugihara2017foot}. However, CoP damping is probably too simple to precisely reproduce human's early-swing lower-leg trajectories \cite{cappellini2006motor}. It is also not powerful enough to balance the robot in very slow stepping frequencies. Therefore, we consider developing more advanced push-off and CoP control algorithms in future work.

\subsection{Limb Behavior}
In our Atlas controller, the two legs function very similarly to the IP model with fixed-length legs during the stance phase. A better control strategy would be to make them virtually compliant (with torque control \cite{capturability2}) to reduce the touch-down impacts during uneven-terrain locomotion. Our 3LP model can also produce slope-walking by rotating the gravity vector which introduces external horizontal forces. This part is left for future work after obtaining a better understanding of human's limb functionalities in slope walking conditions. 

\subsection{Non-Periodic Motions} 
The method presented in this paper is suitable for recovery from external pushes and obtaining periodic gaits. Walking on partial footholds, rubbles, stairs and structured terrains would require powerful perception and planning. The algorithms introduced in \cite{dai2014whole, herzog2015trajectory} can handle these nonlinearities off-line. However, they are probably too slow to give online updates (in milliseconds) needed for extreme disturbance rejection scenarios. For complex environments, our 3LP model is probably too simple to produce feasible gaits, and our time-projection controller cannot handle inequality constraints. However, 3LP mechanics and other intuitions we obtained in this paper on the design of optimization cost functions can help in future to design better MPC controllers. 

\subsection{Phase Timing}
The entire theory of this paper is based on a fixed-timing whereas, in our Atlas controller, we read contact forces and adaptively change the role of hip joints in the robot. In unexpected phase changes, therefore, both the legs do not start their new roles unless the feet establish or de-establish contacts with the ground. An alternative would be to reset the phase once a new contact is detected. For this purpose, our foot-placement laws can be simply truncated or extrapolated. However, in the current implementation, we do not change phase timing as a control strategy to stabilize walking which brings more robustness but introduces a nonlinear control problem \cite{khadiv2016step}. 

\subsection{Torso Rotations}
We saw that bending the torso can produce emergent walking gaits. We also saw that with smaller hip gains, the torso tends to bend forward during walking, although it is commanded to stay upright. At the same time during push-recovery scenarios, the torso may also momentarily rotate back and forth, depending on the hip gain and push strength. In the future, we would like to study the effect of torso momentums on gait stabilization and dynamics. A possible starting point would be to make the torso pendulum free in 3LP model, though with linear dynamics. This may give time-projection rules to control the torso tilt angle as well.

\subsection{Model-Based Control}
The foot-placement strategy in our method stabilizes model violations including heel-toe motions and vertical pelvis excursions. We call our walking gaits imprecise because these violations influence the speed. In other words, even using the 3LP model, we can not exactly determine the final walking speed emerging from for example 10 cm of pelvis shift on the Atlas robot. However, we can still produce a wide range of forward and backward speeds by simulating the system and measuring the velocity. In this regard, although time-projection requires the 3LP model to project errors back in time, our full system is not completely explained. The 3LP model mainly helps to derive more optimal foot-placement rules for heavy-legged robots or fast walking frequencies. 

\subsection{Swing trajectories}
Since we do not use 3LP gaits to produce nominal walking, time-variable swing-hip trajectories are missing in our model. In other words, we directly command the final attack angle to the swing leg and let it reach that point over time. This strategy may produce fast swing motions on the Atlas robot which has relatively light-weight legs and powerful actuators. On heavy-legged robots like COMAN which also has a delay-full control \cite{faraji2015practical}, our strategy ensures a more precise tracking. Alternative methods of shaping the swing trajectories could be tuning the damping terms in the swing-hip controller, or designing time-dependent profiles similar to Fig. \ref{fig::DLQRgain}.D and scaling them by desired step lengths \cite{faraji2014robust}. These improvements could be done in future work. 

This paper presented the concept of time-projection which worked best for the hybrid system of bipedal walking thanks to a smart state transformation which removed hybridity. Our theory produced simple look-up-table control rules that could stabilize a complex humanoid system in extreme perturbations. This simple controller can be applied to other robots as well by finding proper values for the few control parameters involved and recalculation of the foot-placement curves. All the functions used to generate these curves are found in a MATLAB GUI which will be available online at \url{https://biorob.epfl.ch/research/humanoid/walkman}.
\appendices

\section{DLQR for constrained systems}
\label{sec::app_dlqr_const}
For a discrete constrained system of:
\begin{eqnarray}
\nonumber &X[k+1] = A X[k] + B U[k]\\
&CX[k+1] = 0
\label{eqn::simple_system_discrete_appA}
\end{eqnarray}
which has a set-point solution $\bar{X}[k]$ and $\bar{U}[k]$, the error system is defined as:
\begin{eqnarray}
\nonumber &E[k+1] = A E[k] + B \Delta U[k] \\
&CE[k+1] = 0
\label{eqn::simple_system_error dynamics_appA}
\end{eqnarray}
where $E[k] = X[k] - \bar{X}[k]$. Here, regardless of the control strategy, the constraint represented by $C$ should always be satisfied. Consider $A \in \mathbb{R}^{N \times N}$ and $B \in \mathbb{R}^{N \times M}$ and $C \in \mathbb{R}^{P \times N}$. The DLQR optimization problem for this system is:
\begin{eqnarray}
\nonumber & \underset{E[k],\Delta U[k]}{\text{min}} \sum_{k=0}^{\infty} E[k]^TQE[k]+\Delta U[k]^TR\Delta U[k]  \\
&s.t. \left\{ \begin{array}{l} E[k+1] = AE[k]+B\Delta U[k] \\ CE[k+1] = 0 \end{array}
\, \, \right. k\ge0
\label{eqn::simple_system_lqr_appA}
\end{eqnarray}
Assume we find a matrix $\tilde{C} \in \mathbb{R}^{(N-P) \times N}$ that forms a complete basis with $C$. In other words, the matrix $S$ defined by:
\begin{eqnarray}
S = \begin{bmatrix} \tilde{C} \\ C \end{bmatrix}
\label{eqn::simple_system_S}
\end{eqnarray}
has a full rank. Now, we define a new variable $Z[k]=SE[k]$ which will produce the following new DLQR problem:
\begin{eqnarray}
\nonumber & \underset{Z[k],\Delta U[k]}{\text{min}} \sum_{k=0}^{\infty} Z[k]^T \tilde{Q} Z[k]+\Delta U[k]^T \tilde{R} \Delta U[k]  \\
&s.t. \left\{ \begin{array}{l} Z[k+1] = \tilde{A} E[k]+ \tilde{B} \Delta U[k] \\ \tilde{C}Z[k+1] = 0 \end{array}
\, \, \right. k\ge0
\label{eqn::simple_system_lqr_Z_optim}
\end{eqnarray}
where:
\begin{eqnarray}
\nonumber \tilde{Q} = S^{-T} Q S^{-1}, \quad \tilde{R} = R \quad \quad \quad\\
\tilde{A} = S A S^{-1}, \quad \tilde{B} = S B, \quad \tilde{C} = C S^{-1}
\label{eqn::simple_system_lqr_Z}
\end{eqnarray}
Note that:
\begin{eqnarray}
Z[k] = SE[k] = \begin{bmatrix} \tilde{C}E[k] \\ CE[k] \end{bmatrix} = \begin{bmatrix} Y[k] \\ 0 \end{bmatrix}
\label{eqn::simple_system_Z_decompose}
\end{eqnarray}
where $Y[k] \in \mathbb{R}^{N-P}$ is reduced to exclude the constraint. Assuming $P<N$, a full rank $C$ and $M>=P$, we can find $P$ independent components of $\Delta U[k]$ and rewrite the constraint in terms of these components. Without loss of generality, assume these components are the last P components of $\Delta U[k]$, referred to as $\Delta W[k]$ hereafter:
\begin{eqnarray}
\Delta U[k] = \begin{bmatrix} \Delta V[k] \\ \Delta W[k] \end{bmatrix}
\label{eqn::simple_system_U_decompose}
\end{eqnarray}
where $\Delta V[k] \in \mathbb{R}^{M-P}$ and $\Delta W[k] \in \mathbb{R}^{P}$. Consider we decompose matrices $\tilde{Q}$ and $\tilde{R}$ as follows:
\begin{eqnarray}
\tilde{Q} = \begin{bmatrix} \tilde{Q}^{vv} & \tilde{Q}^{vw} \\ \tilde{Q}^{wv} & \tilde{Q}^{ww} \end{bmatrix}, \,\,\,\,
\tilde{R} = \begin{bmatrix} \tilde{R}^{vv} & \tilde{R}^{vw} \\ \tilde{R}^{wv} & \tilde{R}^{ww} \end{bmatrix} 
\label{eqn::simple_system_decompose_QR}
\end{eqnarray}
where the last lower right corner of size $P\times P$ is taken out with an index $^{ww}$. Likewise, system matrices  $\tilde{A}$ and $\tilde{B}$ can be decomposed to:
\begin{eqnarray}
\nonumber \begin{bmatrix} Y[k+1] \\ 0 \end{bmatrix} &=& 
\begin{bmatrix} \tilde{A}^{vv} & \tilde{A}^{vw} \\ \tilde{A}^{wv} & \tilde{A}^{ww} \end{bmatrix}
\begin{bmatrix} Y[k] \\ 0 \end{bmatrix} \\ 
&+& \begin{bmatrix} \tilde{B}^{vv} & \tilde{B}^{vw} \\ \tilde{B}^{wv} & \tilde{B}^{ww} \end{bmatrix}
\begin{bmatrix} \Delta V[k] \\ \Delta W[k] \end{bmatrix}
\label{eqn::simple_system_decompose_AB}
\end{eqnarray}
With such decomposition, we can take $\Delta W[k]$ out of (\ref{eqn::simple_system_decompose_AB}) from the last $P$ rows:
\begin{eqnarray}
\nonumber \Delta W[k] &=& \tilde{G}Y[k] + \tilde{H}\Delta V[k] \\
\nonumber \tilde{G} &=& - (\tilde{B}^{ww})^{-1}\tilde{A}^{wv} \\
\tilde{H} &=& - (\tilde{B}^{ww})^{-1}\tilde{B}^{wv}
\label{eqn::simple_system_W_GH}
\end{eqnarray}
and:
\begin{eqnarray}
\nonumber  Y[k+1] &=& \bar{A}Y[k]+\bar{B}\Delta V[k] \\
\nonumber \bar{A} &=& \tilde{A}^{vv} + \tilde{B}^{vw}  \tilde{G}\\
\bar{B} &=& \tilde{B}^{vv} + \tilde{B}^{vw}  \tilde{H}
\label{eqn::simple_system_lqr_free_appA_system_matrices}
\end{eqnarray}
Now, given that we have resolved the constraint, we can form an equivalent DLQR design in terms of $Y[k]$ and $\Delta V[k]$ by replacing $\Delta W[k]$ in all terms:
\begin{eqnarray}
\nonumber & \underset{Y[k],\Delta V[k]}{\text{min}} \sum_{k=0}^{\infty} \\
\nonumber & Y[k]^T\bar{Q}Y[k]+\Delta V[k]^T\bar{R}\Delta V[k]+2Y[k]^T\bar{N}\Delta V[k] \\
& s.t. \, \, \, Y[k+1] = \bar{A}Y[k]+\bar{B}\Delta V[k]
\, \, \, \, k\ge 0
\label{eqn::simple_system_lqr_free_appA}
\end{eqnarray}
which has a standard DLQR format without constraint. The new cost matrices in (\ref{eqn::simple_system_lqr_free_appA}) are defined as:
\begin{eqnarray}
\nonumber \bar{Q} &=& \tilde{Q}^{vv} + \tilde{G}^T \tilde{R}^{ww} \tilde{G}\\
\nonumber \bar{R} &=& \tilde{R}^{vv} + \tilde{H}^T \tilde{R}^{ww} \tilde{H} + \tilde{R}^{vw} \tilde{H} + \tilde{H}^T \tilde{R}^{wv}\\
\bar{N} &=& \tilde{G}^T ( \tilde{R}^{ww} \tilde{H} + \tilde{R}^{ww}{}^T \tilde{H}^T + \tilde{R}^{vw}{}^T +  \tilde{R}^{wv})
\label{eqn::simple_system_lqr_free_appA_matrices}
\end{eqnarray}
We call the optimal DLQR gain matrix for (\ref{eqn::simple_system_lqr_free_appA}) $\bar{K}$ which acts on $Y[k]=\tilde{C}E[k]$ and produces $\Delta V[k]$. The other part of system input $\Delta W[k]$ can be calculated by (\ref{eqn::simple_system_W_GH}). Overall, the system input is: 
\begin{eqnarray}
\Delta U[k] = \begin{bmatrix} -\bar{K} \\ \tilde{G}-\tilde{H}\bar{K} \end{bmatrix} \tilde{C}E[k]
\label{eqn::simple_system_lqr_free_sol}
\end{eqnarray}
which optimally satisfies the constraint and initial DLQR problem in (\ref{eqn::simple_system_lqr_appA}).

\section{Time-projection for constrained systems}
\label{sec::app_proj_const}

Time-projection controller for a constrained system is derived very similarly to a normal system. Consider all formulations and decompositions of Appendix A. The instantaneous error $e(t)=x(t)-\bar{x}(t)$ in a constrained system would evolve until the next control time by:
\begin{eqnarray}
\hat{E}_t[k+1] = A(\tau) e(t) + B(\tau) \delta \hat{U}_t[k]
\label{eqn::simple_system_xtoX+_appB}
\end{eqnarray}
where $\tau=(k+1)T-t$ and the constraint applies $C\hat{E}_t[k+1] = 0$. Here we assume a constant input $\delta \hat{U}_t[k]$ applied to the system which yields to a predicted error $\hat{E}_t[k+1]$. Remember the hat notation is used to emphasize that these predicted quantities are just calculated at time $t$ and they are not real system variables. Imagine we define $z(t) = Se(t)$, $\tilde{A}_t = S A(\tau) S^{-1}$, $\tilde{B}_t = S B(\tau)$ and perform a decomposition similar to (\ref{eqn::simple_system_decompose_AB}):
\begin{eqnarray}
\nonumber \hat{Z}_t[k+1] &=& \tilde{A}_t z(t) + \tilde{B}_t  \delta \hat{U}_t[k] \\
\nonumber \begin{bmatrix} \hat{Y}_t[k+1] \\ 0 \end{bmatrix} &=& 
\begin{bmatrix} \tilde{A}_t^{v} \\ \tilde{A}_t^{w} \end{bmatrix}
z(t) \\ 
&+& \begin{bmatrix} \tilde{B}_t^{vv} & \tilde{B}_t^{vw} \\ \tilde{B}_t^{wv} & \tilde{B}_t^{ww} \end{bmatrix}
\begin{bmatrix} \delta \hat{V}_t[k] \\ \delta \hat{W}_t[k] \end{bmatrix}
\label{eqn::simple_system_decompose_AcBc}
\end{eqnarray}
which describes system evolution from time $t$ to $(k+1)T$ whereas the constraint in (\ref{eqn::simple_system_lqr_Z_optim}) describes same evolution from $kT$ to $(k+1)T$. Here the subscript $t$ indicates dependency on $t$. Note that the last $P$ elements of $z(t)$ might not be zero in $kT<t<(k+1)T$, but the constraint implies that these last $P$ elements have to be zero at time instances $kT$ and $(k+1)T$. Similar to (\ref{eqn::simple_system_W_GH}), we can take $\delta \hat{W}[k]$ out of the last $P$ equations:
\begin{eqnarray}
\nonumber \delta \hat{W}_t[k] &=& \tilde{G}_t z(t) + \tilde{H}_t \delta \hat{V}_t[k] \\
\nonumber \tilde{G}_t &=& - ({\tilde{B}_t^{ww}})^{-1}\tilde{A}_t^{w} \\
\tilde{H}_t &=& - ({\tilde{B}_t^{ww}})^{-1}\tilde{B}_t^{wv}
\label{eqn::simple_system_W_GcHc}
\end{eqnarray}
and new system matrices would be defined as:
\begin{eqnarray}
\nonumber \hat{Y}_t[k+1] &=& \bar{A}_t z(t) + \bar{B}_t \delta \hat{V}_t[k] \\
\nonumber \bar{A}_t &=& \tilde{A}_t^{v} + \tilde{B}_t^{vw}  \tilde{G}_t\\
\bar{B}_t &=& \tilde{B}_t^{vv} + \tilde{B}_t^{vw}  \tilde{H}_t
\label{eqn::simple_system_lqr_free_AcbarBcbar}
\end{eqnarray}
Note that in (\ref{eqn::simple_system_lqr_free_appA_system_matrices}) and (\ref{eqn::simple_system_lqr_free_AcbarBcbar}), the constraint is resolved. In other words, system matrices are adjusted such that they account for the effect of $W$ inputs which aim at satisfying the constraint. Now, we can consider time-projection for this free system as follows. Imagine an initial reduced state $\hat{Y}_t[k]$ evolves in time by $\delta \hat{V}_t[k]$ according to  (\ref{eqn::simple_system_lqr_free_appA_system_matrices}) and yields to $\hat{Y}_t[k+1]$. Similarly, the current error $z(t)$ evolves in time by $\delta \hat{V}_t[k]$ and yields to the same $\hat{Y}_t[k+1]$ according to (\ref{eqn::simple_system_lqr_free_AcbarBcbar}).  Now, the system of equations in (\ref{eqn::simple_system_solvedU}) can be applied here as well:
\begin{eqnarray}
\begin{bmatrix} \bar{A} & \bar{B}-\bar{B}_t & \cdot\\ K & I & \cdot \\ \cdot & -\tilde{H}_t & I \end{bmatrix} \begin{bmatrix}
\hat{Y}_t[k] \\ \delta \hat{V}_t[k] \\ \delta \hat{W}_t[k] \end{bmatrix} = 
\begin{bmatrix} \bar{A}_t \\ 0 \\ \tilde{G}_t \end{bmatrix} S e(t)
\label{eqn::simple_system_solvedUc}
\end{eqnarray}
where the solution defines $\delta u(t) = \delta \hat{U}_t[k]$ to be applied at time instance $t$.

\section{Push Timing Sensitivity}
\label{sec::app_timing}
\begin{figure*}[]
    \centering
    \includegraphics[trim = 0mm 0mm 0mm 0mm, clip, width=1\textwidth]{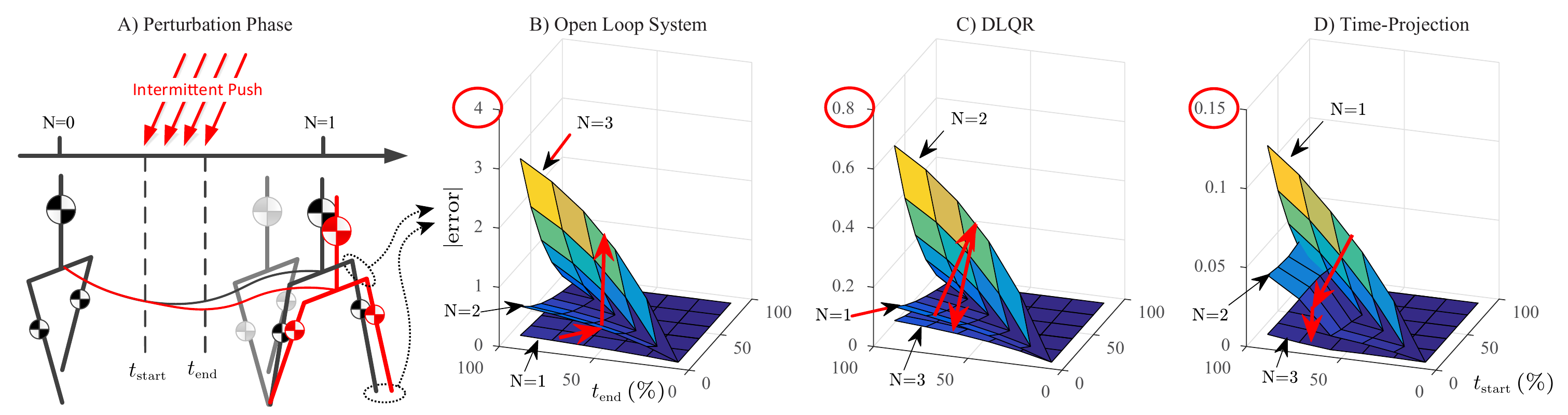}
    \caption{A) Demonstration of an intermittent push that appears shortly during a continuous phase and influences the system. Normal and disturbed trajectories are shown in black and red respectively. B, C, D) recovery performance for the open-loop system, DLQR, and time-projection controllers. In these plots, we note the start and end of the push as the percentage of the phase. Surfaces show error norms, calculated at touch down events over three consecutive steps. The open-loop system is unstable, the DLQR controller overshoots and the time-projection controller reacts immediately while the push is being applied. The time-projection controller is therefore much stronger in rejecting even long-lasting intermittent pushes which span throughout the whole phase.} 
    \label{fig::intermittent}
\end{figure*}

Due to the unstable dynamics of walking, the sensitivity of footstep adjustments with respect to the timing of external pushes is important. To this end, we simulated the open-loop 3LP system and closed-loop DLQR and time-projection controllers with pushes of the same magnitude, but variable timings and durations. The results are shown in Fig. \ref{fig::intermittent} over three consecutive steps. We applied the push between certain times of the phase, shown by percentage on the two horizontal axes. Error surfaces quantify the norm of error with respect to the nominal in-place walking gait, calculated at touch down events and plotted along the vertical axis. Fig. \ref{fig::intermittent} demonstrates that a push of same magnitude and duration might have a more severe effect on the system if applied earlier in the phase. While the open-loop system is unstable, the other two closed-loop systems recover from the push successfully, but with certain dynamics. At the end of the first step, the DLQR controller produces an error almost similar to the open-loop system. However, due to a delayed reaction, it overshoots in the second step. The time-projection controller, however, adjusts the footstep online during the first step to avoid such overshoot. Fig. \ref{fig::intermittent} also indicates that longer pushes have more severe effects, especially if they start earlier in the phase. 

\section{Viable Regions}
\label{sec::app_viability}

\begin{figure}[]
    \centering
    \includegraphics[trim = 0mm 0mm 0mm 0mm, clip, width=0.4\textwidth]{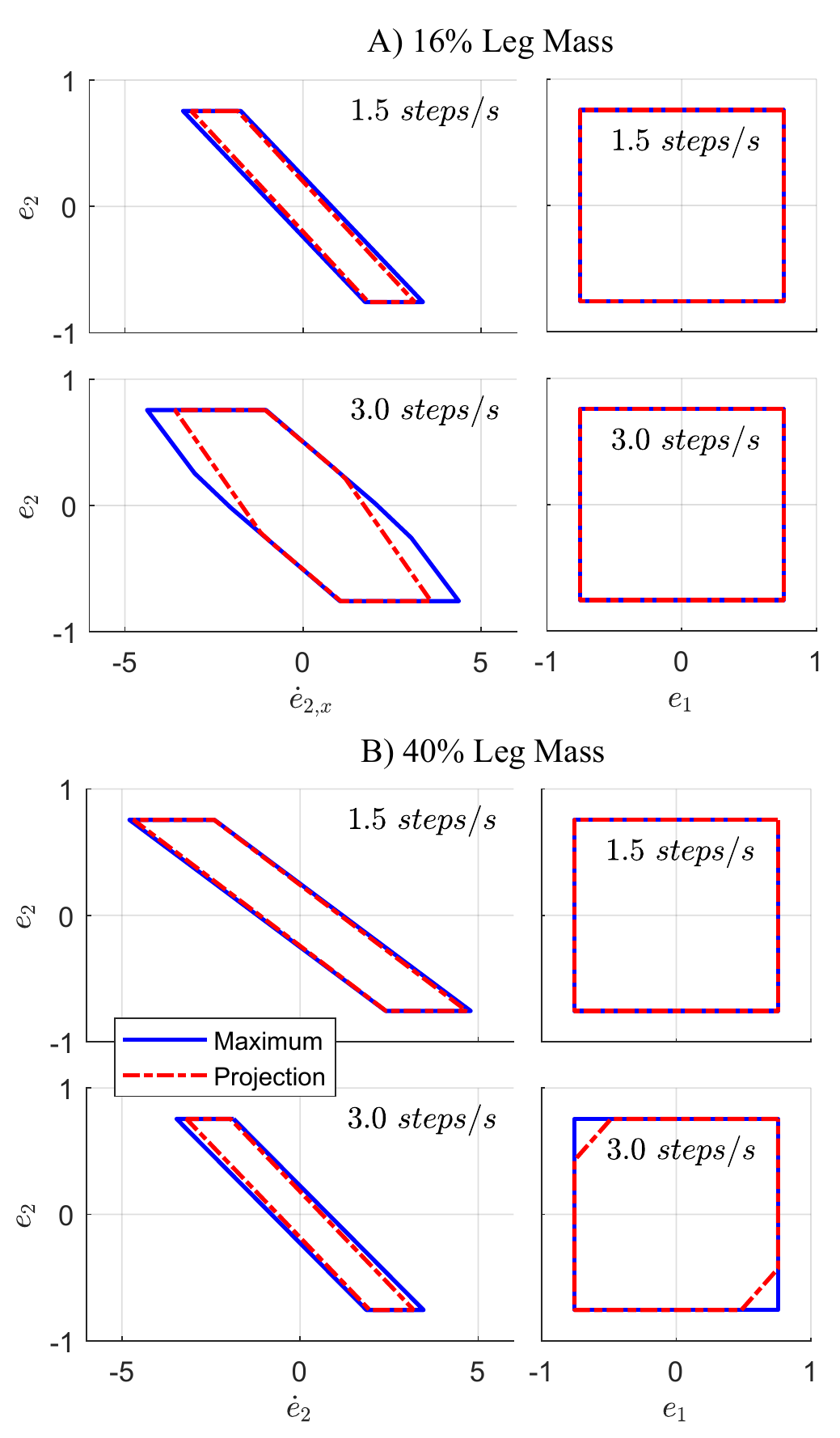}
    \caption{Viable sets for the time-projection controller and the best controller possible (including MPC). These plots are calculated by considering hip torque limits and step length constraints for the Atlas robot. The time-projection controller only reduces the maximum viable set slightly in extreme walking conditions.} 
    \label{fig::regions}
\end{figure}

The goal of this analysis is to find the set of viable states for the time-projection controller \cite{zaytsev2015two} as well as the maximum set for all possible controllers. We apply time-projection to the 3LP model scaled by Atlas dimensions, assume torque limits of 260 Nm in the hip joins, and footstep locations that go up to $0.8l$ away from their hip joints horizontally (which form square regions). Here $l$ denotes the leg length. We divide each phase into five shorter sub-phases where time-projection and arbitrary input profiles (for the maximum viable set) can provide continuous inputs. Our maximum viable set includes all the states which are capturable in six steps, i.e., with feasible arbitrary inputs over these steps. Fig. \ref{fig::regions}.A plots these sets for a normal Atlas robot walking at 1.5 steps/s and 3 steps/s. In slow frequencies, the time-projection controller covers most of the maximum set whereas, in higher frequencies, other controllers can produce more complex profiles of hip torques compared to our piecewise linear profiles. Therefore, swing dynamics can stabilize a slightly larger region of states in higher frequencies. With artificially heavier legs (related to Fig. \ref{fig::massprop}), the maximum viable set shrinks considerably in Fig. \ref{fig::regions}.B, though the time-projection controller still covers most of this maximum set. In conclusion, the plots in Fig. \ref{fig::regions} indicate that although time-projection does not consider inequality constraints, it can stabilize a considerable portion of the maximum set of viable states. Therefore, removing the inequality constraints in time-projection is not a big compromise. In worse cases, the regions plotted in Fig. \ref{fig::regions} can predict failure for a given state, which could then trigger an emergency controller on the real robot.

\section*{Acknowledgment}
This work was funded by the WALK-MAN project (European Community's 7th Framework Programme: FP7-ICT 611832).

\bibliographystyle{IEEEtran}
\bibliography{Biblio}

\ifCLASSOPTIONcaptionsoff
  \newpage
\fi

\end{document}